\pdfoutput=1

\documentclass[11pt]{article}

\usepackage[preprint]{acl}

\usepackage{times}
\usepackage{latexsym}
\usepackage{amsfonts}
\usepackage{amsmath}
\usepackage{verbatim}
\usepackage{placeins}
\usepackage[table,xcdraw]{xcolor}
\usepackage[T1]{fontenc}

\usepackage[utf8]{inputenc}
\usepackage{hyperref}
\usepackage{microtype}

\usepackage{inconsolata}

\usepackage[normalem]{ulem}
\useunder{\uline}{\ul}{}
\usepackage{graphicx}
\usepackage{booktabs}

\usepackage{tcolorbox}

\usepackage{multirow}

\tcbuselibrary{listings,skins,breakable,theorems}

\title{HyperStyler: Low-resource Authorship Style Transfer via Context-aware Style Navigation and Hypernetworks}

\author{
  Jongkyung Shin\textsuperscript{1}$^\dagger$ \;
  Minguk Jeon\textsuperscript{1} \;
  Chanwoo Park\textsuperscript{2} \;
  Chiehyeon Lim\textsuperscript{1,2,3}$^\dagger$
  \\
  \\
    \textsuperscript{1}UNIST\quad
    \textsuperscript{2}POSTECH\quad
    \textsuperscript{3}POSCO Holdings Inc.\quad
  \\
  \small\texttt{ \{shinjk1156, rzbsys\}@unist.ac.kr\quad \{cks1091, chiehyeon.lim\}@postech.ac.kr}
}

\begin{document}
\maketitle
\let\thefootnote\relax\footnotetext{$\dagger$: Corresponding authors.}
\begin{abstract}
Low-resource authorship style transfer (LAST) aims to rewrite text into the style of an arbitrary target author using only a few reference examples while preserving the original meaning. Existing methods often struggle to achieve both high style fidelity and semantic preservation because they compress diverse references into a single static author embedding, which averages out context-dependent stylistic variation, and rely on hidden representations for style control, which entangle style with content. We propose HyperStyler, a novel architecture that decouples LAST into style selection and style realization. Stylo-navigator predicts style coordinates by jointly modeling the source context and target-author references, and Stylo-hypernet realizes them via dynamic parameter modulation instead of hidden-state injection. Our experiments on Reddit, Blog, and News datasets demonstrate that HyperStyler consistently outperforms prior methods including LLM-based approaches and generalizes robustly across domains. Notably, HyperStyler achieves superior performance with as few as 2.4\% additional parameters over T5-large, while being over 1.8× faster than LLMs at inference.
\end{abstract}

\section{Introduction}

Even when writing the same content, individuals exhibit distinctive lexical choices, syntactic structures, and modes of expression \citep{stamatatos2009survey, wang-etal-2023-authorship}. Low-resource authorship style transfer (LAST) aims to rewrite a source text in the style of an arbitrary target author, preserving the original semantics given only a few reference examples \cite{patel2022low}. Unlike traditional style transfer, which is often restricted to authors with massive corpora, LAST extends the scope to everyday writers with a small number of sentences. This shift offers significant practical value, enabling users to efficiently transform drafts into the nuanced voice of any desired author.

Despite its potential, high-fidelity style transfer remains a significant challenge. Initial attempts leverage in-context learning \cite{patel2022low} or inference-time control methods \cite{khan2023learning, horvitz2024paraguide}, but often yield weak stylistic transfer. More recent approaches \cite{liu2024authorship, horvitz-etal-2024-tinystyler} introduce unsupervised style alignment frameworks by constructing pseudo-parallel datasets. However, they still struggle to achieve strong style fidelity and semantic preservation simultaneously.

We can find the first root cause of this challenge in the stylometry literature. According to prior studies, an author's style is not a static template but a multifaceted phenomenon that shifts with topic and register \cite{sapkota-etal-2014-cross, hoover2017microanalysis, grieve2023register}. In the few-shot setting of LAST, this context-dependency introduces a fundamental problem. Since the references provided at inference time each capture the author's style in a specific context, processing all references without explicitly identifying which is most relevant to the source text can result in a style that is either diluted into a generic average or dominated by the most salient reference. Specifically, some methods compressing references into a single static author embedding directly induce mode averaging, while other methods directly feeding all references into the context window lack any mechanism to prioritize contextually relevant references, potentially causing the model to latch onto the most stylistically prominent one.

The second cause lies in how existing methods perform style control in the hidden state space. When stylistic signals are directly injected into hidden states, they become entangled with semantic content, making it difficult to isolate style from semantic content. This is particularly problematic in LAST, where the model must handle the open-ended stylistic variation of unseen authors rather than a fixed set of style categories. Such style-content interference makes it increasingly difficult to precisely realize diverse stylistic variations while preserving the original meaning.

To address these limitations, we propose HyperStyler, a novel architecture that decouples the task into a style selection and a style realization stage via two specialized modules. First, \textit{Stylo-navigator} explicitly predicts style coordinates by considering both the input context and target-author references. Second, \textit{Stylo-hypernet} realizes these coordinates through dynamic parameter modulation, shifting the control mechanism to the parameter space to reduce content-style entanglement and enable fine-grained controllability. Extensive experiments across Reddit, Blog, and News domains demonstrate that HyperStyler consistently outperforms existing baselines including LLM-based approaches, and generalizes robustly across domains. Furthermore, HyperStyler is over 1.8× faster than LLMs at inference, and maintains superior performance even with only a 2.4\% parameter increase over T5-large, highlighting its practical utility for time- and resource-constrained applications.

\section{Related Work}

\subsection{Unsupervised Alignment for LAST} 
Due to the scarcity of parallel data, recent LAST methods commonly follow a two-stage unsupervised alignment framework \citep{krishna-etal-2020-reformulating}. The first stage trains a model to reconstruct the original text from style-neutralized paraphrases, conditioned on either a static author embedding \cite{horvitz-etal-2024-tinystyler} or a set of reference samples \citep{liu2024authorship}. In the second stage, the model is further aligned using filtered pseudo-parallel data. HyperStyler follows this framework but diverges by aligning toward predicted style coordinates rather than a fixed author embedding, enabling the model to account for the author's stylistic variation.

\subsection{Hypernetworks}
Hypernetworks \citep{ha2017hypernetworks} generate parameters of a target model conditioned on an external signal, enabling more flexible modulation than static adapters. They have been primarily used for task- or domain-conditioned adaptation \citep{ivison2023hint, li2024hypernetwork}. However, their use for controlling fine-grained linguistic patterns in open-ended and few-shot settings, where the model must generalize to unseen authors and conditioning signals, remains underexplored. In this paper, we address this gap by conditioning hypernetworks on stylistic coordinates from few-shot references, enabling dynamic style control while reducing content-style entanglement.

\section{HyperStyler}

\subsection{Overview}

\begin{figure}
    \centering
    \includegraphics[width=\columnwidth]{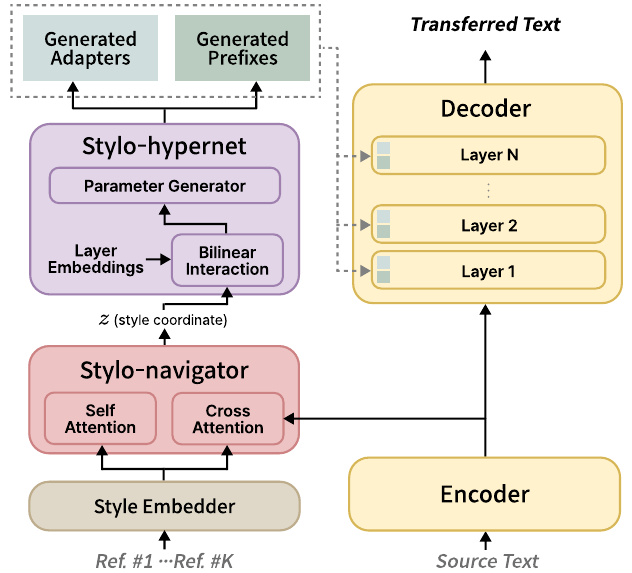}
    \caption{Overall architecture of HyperStyler}
    \label{fig:figure1}
\end{figure}

Given a source text $x$ and a set of references $R=\{r_i\}^K_{i=1}$ written by a target author, our goal is to generate $y$ that matches the target author's writing style while preserving the semantics of $x$. We explicitly decompose this process into two subtasks: (1) style selection, which infers a target style from the references and source context, and (2) stylistic realization, which rewrites $x$ in the target style without altering its meaning. 

As illustrated in Figure \ref{fig:figure1}, HyperStyler implements these subtasks via two modules attached to an encoder--decoder paraphraser. This architecture decouples content encoding in the encoder from stylistic realization in the decoder, aligning with established practices in style transfer \cite{krishna-etal-2020-reformulating, lee2021enhancing, zhao-etal-2024-sc2}. First, the Stylo-navigator predicts a style coordinate $z$ from $x$ and $R$. To ensure parameter efficiency, we reuse the backbone encoder representations $H_{\text{enc}}=\text{Enc}(x)$ as the contextual signal for $x$ without an extra encoder. Second, the Stylo-hypernet generates parameter modulations conditioned on $z$, which are applied to the decoder as key/value prefixes in the attention layers and low-rank weight updates in the feed-forward networks (FFNs).

\subsection{Stylo-navigator}
We define a stylistic coordinate $z$ in the style embedding space using STYLE embedder \citep{wegmann-etal-2022-author} trained to capture content-independent stylistic representations. This reduces the influence of content semantics from the reference sentences on the control signal, encouraging $z$ to primarily reflect stylistic characteristics. Each reference sentence $r_i$ is mapped to a style embedding $s_i$, yielding a set of reference embeddings $S \in \mathbb{R}^{K\times d}$. 

The Stylo-navigator predicts a stylistic coordinate $z$ by attending reference weights conditioned on the source context via two parallel attention mechanisms. We apply self-attention over $S$ to capture inter-reference stylistic patterns that characterize the author's uniqueness, producing $\tilde{S}=\mathrm{SelfAttn}(S)$. In parallel, cross-attention is applied with $H_{\text{enc}}$ as queries and $S$ as keys and values, where each token-level hidden state attends to the entire reference set for fine-grained, context-dependent style selection. The token-specific results are aggregated via mean pooling to form a context-aware style query $q \in \mathbb{R}^{d}$:
\begin{equation}q = \mathrm{MeanPool}\big(\mathrm{CrossAttn}(H_{\text{enc}}, S, S)\big).
\end{equation}
We then compute a scaled dot product between $q$ and each $\tilde{s_i} \in \mathbb{R}^{d}$ to obtain the contribution weight $\alpha_i$ of each reference:
\begin{equation}
\alpha_i = \frac{\exp(\bar{q} \cdot \bar{\tilde{s_i}} / \sqrt{d})}{\sum_{k=1}^{K} \exp(\bar{q} \cdot \bar{\tilde{s_k}} / \sqrt{d})},
\end{equation}
where $\bar{q}$ and $\bar{\tilde{s}}_i$ are layer-normalized for stability. The stylistic coordinate $z$ is obtained as the weighted sum of the reference embeddings:
\begin{equation}
z = \sum_{i=1}^{K} \alpha_i s_i.
\end{equation}

Note that $z$ lies within the \textsc{Style} space, not in a newly defined space. Because the weights $\alpha_i$ are conditioned on the source context, $z$ varies with the context even for the same author. As an interpolation rather than a selection, $z$ can also reach coordinates between individual references.

\subsection{Stylo-hypernet}
Stylo-hypernet dynamically modulates the decoder conditioned on the stylistic coordinate $z$. Prior analyses suggest that transformer layers contribute differently to generation \cite{langedijk2024decoderlens, alshomary2025layered}, implying that stylistic realization may be inherently layer-dependent. Motivated by this observation and inspired by \citet{ivison2023hint}, we introduce learnable layer embeddings and modulate them with $z$ to construct layer-specific style signals. Concretely, we compute a style-dependent offset for each modulation target and add it to the corresponding layer embedding via a residual connection to preserve layer identity. The resulting layer-specific signals are then used to generate modulation parameters for the decoder.

\paragraph{Style-conditioned Layer Embeddings.}
Let $E^{(t)}=[e_1;\dots;e_{N_t}]\in\mathbb{R}^{N_t\times d_e}$ be a learnable embedding table associated with a modulation target $t$ (e.g., cross-attention prefix keys). Here, $N_t$ is the number of embeddings for target $t$, and each row $e_j$ corresponds to a distinct modulation target indexed by a tuple (layer, type, position). Prefix embeddings are indexed by (layer, key/value, prefix position) and adapter embeddings are indexed only by projection type (layer, up/down).

We compute compatibility scores between $z$ and each layer embedding through a multi-head bilinear interaction:
\begin{equation}
b_j^{(h)} = (W_e \bar{e}_j)^{(h)\top} (W_s \bar{z})^{(h)},
\label{eq4}
\end{equation}
where $(\cdot)^{(h)}$ is the $h$-th head, $W_e$ and $W_s$ are trainable projection matrices, and $\bar{e}_j$ and $\bar{z}$ denote layer-normalized vectors. Each $b_j^{(h)}$ is a style-dependent and embedding-specific signal that determines the relative contribution of the corresponding subspace of $z$ to embedding $j$. We then form the offset $o_j$ by 
projecting $\bar{z}$ with trainable matrix $W_v$, scaling each 
head with its corresponding score, and concatenating all $N_h$ heads:
\begin{equation}
o_j = \big[ b_j^{(1)}(W_v \bar{z})^{(1)} ; \dots ; b_j^{(N_h)}(W_v \bar{z})^{(N_h)} \big].
\end{equation}
Finally, we project the offset $o_j$ with trainable matrix $W_o$ 
and apply layer normalization to obtain a stable style-conditioned 
update $\Delta e_j$, which is added to the original embedding 
$e_j$:
\begin{equation}
\Delta e_j = \mathrm{LN}(W_o o_j), \quad 
\tilde{e}_j = e_j + \Delta e_j.
\end{equation}

\paragraph{Generating Modulation Parameters.}
We map the style-conditioned layer embeddings $\tilde E^{(t)}$ to modulation parameters using two-layer MLPs. We use dedicated generators for different modulation targets, as each target operates on distinct parameter spaces with different dimensionalities and functional roles. Here, $d_{\text{model}}$ denotes the hidden dimensionality of the underlying model.

\textbf{Cross-attention prefixes} modulate how the decoder references the source context during generation \cite{li2021prefix}. We generate key and value prefix vectors using two independent MLPs. The generated vectors are grouped into length-$p$ prefixes per layer,
$P^{\ell}_{K}, P^{\ell}_{V} \in \mathbb{R}^{p \times d_{\text{model}}}$, and concatenated to the original keys and values along the sequence dimension, $K^{'\ell}=[P^{\ell}_{K};K^{\ell}]$ and $V^{'\ell}=[P^{\ell}_{V};V^{\ell}]$.

\textbf{Low-rank adapters} target FFN layers, which have been shown to store substantial linguistic information \cite{geva2021transformer, geva2022transformer}, making them well-suited for controlling surface realization such as lexical choice and syntactic patterns. For each layer $\ell$, we generate low-rank down- and up-projection weights using separate MLPs, $W^{\ell}_{down}\in\mathbb{R}^{d_{\text{model}}\times r}$ and $W^{\ell}_{up}\in\mathbb{R}^{r\times d_{\text{model}}}$. The resulting branch is added to the FFN output:
\begin{equation}
h^{l}_{\text{out}}=\mathrm{FFN}^{l}(h^{l}_{\text{in}})+\sigma(h^{l}_{\text{in}}W^{\ell}_{down})W^{\ell}_{up},
\label{eq_lora}
\end{equation}
where $h_{\text{in}}$ and $\sigma(\cdot)$ denote the FFN input hidden states and the activation function, respectively. In our implementation, each MLP outputs a vector of length $r\cdot d_{\text{model}}$, which is reshaped into the corresponding low-rank matrix for each layer.

\subsection{Training Procedure}
Due to the lack of a parallel dataset for LAST, we adopt an unsupervised training setting, and the overall training procedure consists of three stages.

\paragraph{Stage 1: Training the Underlying Model.}
For each author, we collect an author-specific corpus consisting of a sentence set $X=\{x_i\}_{i=1}^{K}$. Using a pretrained paraphraser, we generate a synthetic paraphrase $x'_i$ for each sentence $x_i$, thereby constructing synthetic pairs $\{(x_i, x'_i)\}_{i=1}^{K}$. To mitigate stylistic bias inherited from the pretrained paraphraser and to improve both diversity and semantic preservation, we train the underlying model with a bidirectional reconstruction objective over $(x_i \leftrightarrow x'_i)$ pairs \cite{sjoblom-etal-2020-paraphrase, ma2021collaborative}. The goal of this stage is not to acquire any specific style, but to establish a semantically reliable paraphrasing backbone.

\paragraph{Stage 2: Training the Stylo-navigator and Stylo-hypernet.}
We freeze the underlying paraphraser and integrate it with the Stylo-navigator and Stylo-hypernet to be trained simultaneously through an unsupervised reconstruction task. For each author, the reference set $R=X$ is embedded into the style embeddings $S \in \mathbb{R}^{K \times d}$. The Stylo-navigator is trained to identify the stylistic target within $S$ that best matches $x_i$ given the source context $x'_i$. We use the index $i$ of the target sentence as a ground-truth label and minimize the negative log-likelihood of the predicted selection probabilities $\alpha$:
\begin{equation}
\mathcal{L}_{\text{nav}} = - \sum \log \alpha_i.
\end{equation}
To prevent the navigator's prediction errors from propagating to the Stylo-hypernet, we use teacher-forced style conditioning, where the Stylo-hypernet is conditioned on the ground-truth style embedding $s_i$ (obtained by encoding $x_i$ with the STYLE embedder) rather than the predicted coordinate $z$. This isolates hypernetwork optimization from navigator errors and allows the Stylo-hypernet to focus on learning precise parameter modulation. The Stylo-hypernet is optimized to maximize the reconstruction likelihood of $x_i$:
\begin{equation}
\mathcal{L}_{\text{hypernet}} = - \sum \log p(x_{i}\mid x'_i, s_i).
\end{equation}

\paragraph{Stage 3: Unsupervised Alignment Training.}
Finally, we optimize the model for style transfer beyond reconstruction, conditioning the Stylo-hypernet on the stylistic coordinate $z$ from the Stylo-navigator. Since parallel data between source and target authors is unavailable, we construct a high-quality parallel dataset via self-distillation \cite{zhang2019your}. Specifically, we generate style-transferred outputs using the Stage 2 model and filter them following the \textit{Rerank and Filtering} procedure from \citet{horvitz-etal-2024-tinystyler}. Unlike prior work that uses a mean-pooled reference embedding, we use predicted $z$ to evaluate style fidelity during filtering. Using the pseudo-parallel data, we jointly train the Stylo-navigator and Stylo-hypernet.

\begin{table*}[htb]
\centering
\resizebox{\textwidth}{!}{%
\begin{tabular}{@{}lcccccccccccc@{}}
\toprule
\multirow{2}{*}{Method} & \multicolumn{4}{c|}{\textbf{Reddit}} & \multicolumn{4}{c|}{\textbf{Blog}} & \multicolumn{4}{c}{\textbf{News}} \\
 & $\textsc{Away}$ & $\textsc{Towards}$ & $\textsc{Sim}$ & \multicolumn{1}{c|}{$\textsc{Joint}$} & $\textsc{Away}$ & $\textsc{Towards}$ & $\textsc{Sim}$ & \multicolumn{1}{c|}{$\textsc{Joint}$} & $\textsc{Away}$ & $\textsc{Towards}$ & $\textsc{Sim}$ & $\textsc{Joint}$ \\ \midrule
STYLL(Qwen2.5-7B) & 0.811 & 0.063 & 0.428 & \multicolumn{1}{c|}{0.208} & 0.812 & 0.071 & 0.408 & \multicolumn{1}{c|}{0.179} & 0.739 & 0.009 & 0.479 & 0.061 \\
GPT4-turbo & 0.814 & 0.081 & 0.702 & \multicolumn{1}{c|}{0.314} & 0.896 & 0.128 & 0.713 & \multicolumn{1}{c|}{0.331} & 0.677 & 0.063 & 0.860 & 0.290 \\
GPT5-mini & 0.856 & 0.082 & 0.728 & \multicolumn{1}{c|}{0.332} & 0.885 & 0.126 & 0.687 & \multicolumn{1}{c|}{0.333} & 0.719 & 0.093 & 0.759 & 0.336 \\
GPT5.4 & 0.918 & 0.117 & 0.597 & \multicolumn{1}{c|}{0.359} & 0.958 & 0.093 & 0.526 & \multicolumn{1}{c|}{0.276} & 0.848 & 0.071 & 0.718 & 0.273 \\
Llama3.1-8B-Instruct & 0.756 & 0.135 & 0.587 & \multicolumn{1}{c|}{0.390} & 0.875 & 0.173 & 0.539 & \multicolumn{1}{c|}{0.407} & 0.819 & 0.107 & 0.493 & 0.281 \\ \midrule
ParaGuide$_{\lambda=200}$ & 0.763 & 0.053 & 0.598 & \multicolumn{1}{c|}{0.235} & 0.661 & 0.069 & 0.719 & \multicolumn{1}{c|}{0.301} & 0.595 & 0.038 & 0.545 & 0.187 \\
ParaGuide$_{\lambda=2500}$ & 0.853 & 0.067 & 0.450 & \multicolumn{1}{c|}{0.240} & 0.736 & 0.100 & 0.627 & \multicolumn{1}{c|}{0.341} & 0.515 & 0.026 & 0.685 & 0.164 \\
StyleMC & 0.658 & 0.036 & 0.450 & \multicolumn{1}{c|}{0.154} & 0.565 & 0.063 & 0.439 & \multicolumn{1}{c|}{0.195} & 0.403 & 0.029 & 0.574 & 0.153 \\ \midrule
ASTRAPOP & 0.578 & 0.027 & 0.728 & \multicolumn{1}{c|}{0.171} & 0.997 & 0.255 & 0.014 & \multicolumn{1}{c|}{0.060} & 0.840 & 0.060 & 0.170 & 0.139 \\
ASTRAPOP$_{\text{JOINT}}$ & 0.620 & 0.029 & 0.695 & \multicolumn{1}{c|}{0.173} & 0.840 & 0.070 & 0.171 & \multicolumn{1}{c|}{0.139} & 0.576 & 0.082 & 0.713 & 0.319 \\
TinyStyler$_{\text{REC}}$ & 0.897 & 0.144 & 0.352 & \multicolumn{1}{c|}{0.323} & 0.791 & 0.134 & 0.603 & \multicolumn{1}{c|}{0.379} & 0.605 & 0.053 & 0.582 & 0.223 \\
TinyStyler$_{\text{REC,RERANK(5)}}$ & 0.888 & 0.141 & 0.506 & \multicolumn{1}{c|}{0.387} & 0.793 & 0.135 & 0.721 & \multicolumn{1}{c|}{0.421} & 0.598 & 0.054 & 0.691 & 0.252 \\
TinyStyler & 0.860 & 0.122 & 0.626 & \multicolumn{1}{c|}{0.399} & 0.743 & 0.129 & 0.786 & \multicolumn{1}{c|}{0.434} & 0.541 & 0.057 & 0.797 & 0.278 \\
TinyStyler$_{\text{RERANK(5)}}$ & 0.859 & 0.122 & 0.730 & \multicolumn{1}{c|}{0.436} & 0.756 & 0.128 & 0.844 & \multicolumn{1}{c|}{0.452} & 0.561 & 0.057 & 0.843 & 0.294 \\ \midrule
HyperStyler$_{\text{REC}}$ & 0.806 & 0.155 & 0.475 & \multicolumn{1}{c|}{0.378} & 0.676 & 0.187 & 0.684 & \multicolumn{1}{c|}{0.477} & 0.581 & 0.110 & 0.615 & 0.355 \\
HyperStyler$_{\text{REC,RERANK(5)}}$ & 0.800 & 0.152 & 0.705 & \multicolumn{1}{c|}{0.460} & 0.692 & 0.189 & 0.854 & \multicolumn{1}{c|}{0.537} & 0.587 & 0.101 & 0.802 & {\uline{0.399}} \\
HyperStyler & 0.818 & 0.152 & 0.578 & \multicolumn{1}{c|}{\textbf{0.418}} & 0.731 & 0.183 & 0.701 & \multicolumn{1}{c|}{\textbf{0.489}} & 0.571 & 0.098 & 0.678 & \textbf{0.370} \\
HyperStyler$_{\text{RERANK(5)}}$ & 0.815 & 0.147 & 0.791 & \multicolumn{1}{c|}{{\uline{0.485}}} & 0.736 & 0.183 & 0.864 & \multicolumn{1}{c|}{{\uline{0.538}}} & 0.585 & 0.083 & 0.865 & 0.372 \\ \bottomrule
\end{tabular}%
}
\caption{Performance comparison results on three datasets. For the Reddit dataset, we report the average performance across three splits. The highest \textsc{Joint} scores without and with reranking are \textbf{bolded} and \underline{underlined}, respectively.}
\label{table1}
\end{table*}

\section{Experiments}

\subsection{Experimental Setup}
\paragraph{Datasets.}
We conduct experiments on three datasets with distinct genres: \textbf{Reddit} \cite{khan-etal-2021-deep}, \textbf{Blog} \cite{schler2006effects}, and \textbf{News} (All-the-news). 
Following prior works for LAST \cite{patel2022low, horvitz2024paraguide, horvitz-etal-2024-tinystyler}, we segment each author’s corpus into sentences and randomly sample 10 sentences per author. We filter out any samples exceeding 60 tokens and split the data by author into training, validation, and test sets with a 0.9/0.05/0.05 ratio. More details of the datasets are described in the Appendix \ref{data_description}.

We use three evaluation sets of \textbf{Reddit} from \citet{patel2022low}: \textit{Random}, \textit{Single}, and \textit{Diverse}. Each split comprises 15 source and 15 target authors with 16 samples each, totaling 225 transfer directions and 3,600 transformations. We apply the same configuration to \textbf{Blog} and \textbf{News} by randomly selecting hold-out authors from the test dataset.

\paragraph{Evaluation Metrics.}
To ensure fair comparison, we follow the evaluation protocol established in prior LAST studies \cite{patel2022low, horvitz-etal-2024-tinystyler}. \textsc{Away} and \textsc{Towards} measure the degree to which the set of transferred texts moves away from the source author's style and toward the target author's style, respectively. These metrics are computed using a held-out UAR embedder \cite{rivera-soto-etal-2021-learning} trained via contrastive learning for authorship verification. To evaluate semantic preservation, we use the Mutual Implication Score \cite{babakov-etal-2022-large} as \textsc{Sim}. Finally, the \textsc{Joint} score summarizes overall performance of style transfer: $\textsc{Joint}=G(G(\textsc{Towards}, \textsc{Away}), \textsc{Sim})$, where $G(\cdot)$ denotes the geometric mean. (see details in \ref{Metric_Formula})

\paragraph{Baselines.}
We compare against baselines across three categories for the LAST task.

\textbf{In-context learning} methods include STYLL \cite{patel2022low} with Qwen2.5, and models prompted with the instructions from \citet{horvitz-etal-2024-tinystyler}, including Llama3.1, GPT4-turbo, and the reasoning enabled GPT5-mini and GPT5.4.

\textbf{Inference-time control} methods include ParaGuide \cite{horvitz2024paraguide}, a diffusion-based model, and StyleMC \cite{khan2023learning}, which performs Metropolis-Hastings sampling guided by a future regressor.

\textbf{Unsupervised alignment} methods include TinyStyler \cite{horvitz-etal-2024-tinystyler}, which conditions on a mean-pooled style embedding for references and utilizes self-distillation, and ASTRAPOP \cite{liu2024authorship}, a policy optimization conditioning on the entire reference sentences. Beyond its original length-based reward, we also train ASTRAPOP using the \textsc{Joint} as a reward.

\paragraph{Implementation Details.}
To ensure fair comparison, we apply two principles: (1) we unify the backbone to T5-large \cite{raffel2020exploring} across all trainable baselines so that performance reflects methodological rather than capacity differences, and (2) for baselines that require a style guide, we provide a mean-pooled style embedding rather than the UAR embedding used for evaluation to prevent baselines from directly optimizing the evaluation metric. We utilize off-the-shelf paraphrasing PEGASUS \cite{zhang2019pegasus}, following \citet{horvitz-etal-2024-tinystyler}, and set the adapter rank to 32 and the prefix length to 5. Following TinyStyler, we apply \textit{reranking} \cite{suzgun-etal-2022-prompt} at inference time, but use the predicted $z$ as the target style instead of a mean-pooled embedding. Other details are provided in the Appendix \ref{implementation_details}. Our implementation code for HyperStyler is available at \url{https://github.com/JK-SHIN-PG/HyperStyler}.

\begin{table*}[t]
\centering
\resizebox{\textwidth}{!}{%
\begin{tabular}{@{}c|cccccccccccccccc@{}}
\toprule
Model & \textsc{Away} & \textsc{Towards} & \textsc{Sim} & \multicolumn{1}{c|}{\textsc{Joint}} & \textsc{Away} & \textsc{Towards} & \textsc{Sim} & \multicolumn{1}{c|}{\textsc{Joint}} & \textsc{Away} & \textsc{Towards} & \textsc{Sim} & \multicolumn{1}{c|}{\textsc{Joint}} & \textsc{Away} & \textsc{Towards} & \textsc{Sim} & \textsc{Joint} \\ \midrule \midrule
 & \multicolumn{16}{c}{Train: Reddit} \\ \cmidrule(l){2-17} 
 & \multicolumn{4}{c|}{Reddit $\rightarrow$ Blog} & \multicolumn{4}{c|}{Reddit $\rightarrow$ News} & \multicolumn{4}{c|}{Blog $\rightarrow$ Reddit} & \multicolumn{4}{c}{News $\rightarrow$ Reddit} \\ \midrule
TinyStyler & 0.768 & 0.175 & 0.653 & \multicolumn{1}{c|}{0.477} & 0.716 & 0.123 & 0.621 & \multicolumn{1}{c|}{0.412} & 0.672 & 0.093 & 0.797 & \multicolumn{1}{c|}{0.393} & 0.428 & 0.037 & 0.843 & \cellcolor[HTML]{FFCCC9}0.231 \\
HyperStyler & 0.757 & 0.239 & 0.601 & \multicolumn{1}{c|}{\textbf{0.499}} & 0.718 & 0.182 & 0.567 & \multicolumn{1}{c|}{\textbf{0.446}} & 0.755 & 0.209 & 0.604 & \multicolumn{1}{c|}{\textbf{0.481}} & 0.660 & 0.197 & 0.614 & \textbf{0.459} \\ \midrule \midrule
 & \multicolumn{16}{c}{Train: Blog} \\ \cmidrule(l){2-17} 
 & \multicolumn{4}{c|}{Blog $\rightarrow$ News} & \multicolumn{4}{c|}{Blog $\rightarrow$ Reddit} & \multicolumn{4}{c|}{News $\rightarrow$ Blog} & \multicolumn{4}{c}{Reddit $\rightarrow$ Blog} \\ \midrule 
TinyStyler & 0.632 & 0.103 & 0.759 & \multicolumn{1}{c|}{0.399} & 0.678 & 0.094 & 0.794 & \multicolumn{1}{c|}{\textbf{0.400}} & 0.469 & 0.030 & 0.835 & \multicolumn{1}{c|}{\cellcolor[HTML]{FFCCC9}0.224} & 0.765 & 0.176 & 0.662 & 0.479 \\
HyperStyler & 0.630 & 0.117 & 0.720 & \multicolumn{1}{c|}{\textbf{0.411}} & 0.618 & 0.104 & 0.718 & \multicolumn{1}{c|}{0.397} & 0.596 & 0.154 & 0.718 & \multicolumn{1}{c|}{\textbf{0.442}} & 0.762 & 0.252 & 0.639 & \textbf{0.523} \\ \midrule \midrule
\textbf{} & \multicolumn{16}{c}{Train: News} \\ \cmidrule(l){2-17} 
 & \multicolumn{4}{c|}{News $\rightarrow$ Blog} & \multicolumn{4}{c|}{News $\rightarrow$ Reddit} & \multicolumn{4}{c|}{Blog $\rightarrow$ News} & \multicolumn{4}{c}{Reddit $\rightarrow$ News} \\ \midrule
TinyStyler & 0.470 & 0.031 & 0.837 & \multicolumn{1}{c|}{\cellcolor[HTML]{FFCCC9}0.230} & 0.426 & 0.037 & 0.846 & \multicolumn{1}{c|}{\cellcolor[HTML]{FFCCC9}0.237} & 0.633 & 0.101 & 0.754 & \multicolumn{1}{c|}{0.396} & 0.713 & 0.123 & 0.613 & 0.412 \\
HyperStyler & 0.516 & 0.100 & 0.737 & \multicolumn{1}{c|}{\textbf{0.391}} & 0.444 & 0.066 & 0.725 & \multicolumn{1}{c|}{\textbf{0.315}} & 0.714 & 0.193 & 0.593 & \multicolumn{1}{c|}{\textbf{0.454}} & 0.779 & 0.250 & 0.507 & \textbf{0.468} \\ \bottomrule
\end{tabular}%
}
\caption{Cross-domain authorship style transfer performance across three datasets. \textit{Source $\rightarrow$ Target} indicates the transformation of texts from a source domain author's style to a target domain author's style. The highest \textsc{Joint} scores for each pair are in bold; values below 0.3 shaded in \colorbox[HTML]{FFCCC9}{red}. We report results without \textit{reranking}.}
\label{table2}
\end{table*}

\begin{table*}[t]
\centering
\resizebox{0.90\textwidth}{!}{%
\begin{tabular}{@{}c|cccccccccccc@{}}
\toprule
Model & \textsc{Away} & \textsc{Towards} & \textsc{Sim} & \multicolumn{1}{c|}{\textsc{Joint}} & \textsc{Away} & \textsc{Towards} & \textsc{Sim} & \multicolumn{1}{c|}{\textsc{Joint}} & \textsc{Away} & \textsc{Towards} & \textsc{Sim} & \textsc{Joint} \\ \midrule \midrule
 & \multicolumn{12}{c}{Train: Reddit} \\ \cmidrule(l){2-13} 
 & \multicolumn{4}{c|}{Reddit $\rightarrow$ Reddit} & \multicolumn{4}{c|}{Blog $\rightarrow$ Blog} & \multicolumn{4}{c}{News $\rightarrow$ News} \\ \midrule 
TinyStyler & 0.860 & 0.122 & 0.626 & \multicolumn{1}{c|}{\cellcolor[HTML]{FFCCC9}0.399} & 0.776 & 0.111 & 0.733 & \multicolumn{1}{c|}{0.388} & 0.600 & 0.017 & 0.762 & 0.123 \\
HyperStyler & 0.818 & 0.152 & 0.578 & \multicolumn{1}{c|}{\textbf{0.418}} & 0.779 & 0.172 & 0.669 & \multicolumn{1}{c|}{\cellcolor[HTML]{FFFC9E}\textbf{0.476}} & 0.601 & 0.045 & 0.724 & \textbf{0.249} \\ \midrule \midrule
 & \multicolumn{12}{c}{Train: Blog} \\ \cmidrule(l){2-13} 
 & \multicolumn{4}{c|}{Reddit $\rightarrow$ Reddit} & \multicolumn{4}{c|}{Blog $\rightarrow$ Blog} & \multicolumn{4}{c}{News $\rightarrow$ News} \\ \midrule
TinyStyler & 0.828 & 0.061 & 0.711 & \multicolumn{1}{c|}{0.276} & 0.743 & 0.129 & 0.786 & \multicolumn{1}{c|}{\cellcolor[HTML]{FFCCC9}0.434} & 0.538 & 0.034 & 0.800 & 0.216 \\
HyperStyler & 0.787 & 0.073 & 0.641 & \multicolumn{1}{c|}{\textbf{0.304}} & 0.731 & 0.183 & 0.701 & \multicolumn{1}{c|}{\cellcolor[HTML]{FFFC9E}\textbf{0.489}} & 0.556 & 0.057 & 0.746 & \textbf{0.284} \\ \midrule \midrule
 & \multicolumn{12}{c}{Train: News} \\ \cmidrule(l){2-13} 
 & \multicolumn{4}{c|}{Reddit $\rightarrow$ Reddit} & \multicolumn{4}{c|}{Blog $\rightarrow$ Blog} & \multicolumn{4}{c}{News $\rightarrow$ News} \\ \midrule
TinyStyler & 0.718 & 0.029 & 0.726 & \multicolumn{1}{c|}{0.182} & 0.673 & 0.046 & 0.780 & \multicolumn{1}{c|}{0.243} & 0.541 & 0.057 & 0.797 & \cellcolor[HTML]{FFCCC9}0.278 \\
HyperStyler & 0.766 & 0.036 & 0.614 & \multicolumn{1}{c|}{\textbf{0.185}} & 0.694 & 0.120 & 0.712 & \multicolumn{1}{c|}{\cellcolor[HTML]{FFFC9E}\textbf{0.428}} & 0.578 & 0.094 & 0.683 & \textbf{0.365} \\ \bottomrule
\end{tabular}%
}
\caption{In-domain authorship style transfer performance across three datasets. The highest \textsc{Joint} scores are in \textbf{bold} for each pair. Best results are shaded for each training domain: \colorbox[HTML]{FFCCC9}{red} for TinyStyler and \colorbox[HTML]{FFFC9E}{yellow} for HyperStyler.}
\label{table3}

\end{table*}

\subsection{Results} \label{sec:results}

\paragraph{Overall Performance.}
Style transfer requires simultaneously achieving high style fidelity and semantic preservation, as a model biased toward preservation fails to transfer style, while one biased toward style fidelity risks distorting meaning \cite{fu2018style, horvitz-etal-2024-tinystyler}. As shown in Table~\ref{table1}, HyperStyler strikes the best balance between the two objectives, improving \textsc{Towards} while maintaining competitive \textsc{Sim} scores, resulting in the highest \textsc{Joint} scores across all three domains. Further improvements are observed when reranking is applied.

Most baselines have relatively low performance on the News, as news articles are more formal and exhibit lower stylistic variability, making it more difficult to capture distinctive author-specific styles~\cite{eder-etal-2021-acquiring, wang-riddell-2022-cctaa} (Table~\ref{tab:intra_inter_author_distance}). Despite this, HyperStyler outperforms all baselines, including LLMs. Additional experimental results including qualitative analysis are provided in Appendix \ref{additional_experimental_results}.

\paragraph{Generalization Capability.} 

We compare the generalization capability of HyperStyler with TinyStyler, the strongest baseline in our experiments. As shown in Table~\ref{table2}, TinyStyler consistently degrades when transferring from News to Reddit and Blog, where the inter-author distances to the target authors are approximately 2.04$\times$ and 1.68$\times$ larger than those in the in-domain setting, respectively (Figure~\ref{fig:cross_domain_inter_author}). Given these larger distances and the high style variation in Blog and Reddit, this degradation suggests that a single mean embedding fails to provide sufficiently fine-grained style signals for such large stylistic shifts. In contrast, HyperStyler achieves consistently strong performance across most domain pairs, as its context-aware style selection and parameter modulation enable more precise stylistic adaptation across diverse domains. 

We also evaluate in-domain authorship style transfer under out-of-domain training. As shown in Table~\ref{table3}, TinyStyler performs well when trained and evaluated within the same domain, but its performance degrades substantially when applied to other domains. Since the training domain shapes the range and granularity of styles a model learns, and each target domain may require a different level of stylistic precision, some degree of degradation under domain shift is expected. Nevertheless, HyperStyler exhibits relatively limited degradation. Notably, HyperStyler trained only on News achieves Blog$\rightarrow$Blog transfer performance close to that of TinyStyler trained directly on Blog, highlighting HyperStyler's robust generalization capability.

\subsection{Analysis and Ablation Study}
\paragraph{Capturing Context-dependent Style Variation.}

\begin{figure}[!h]
    \centering
\includegraphics[width=\columnwidth]{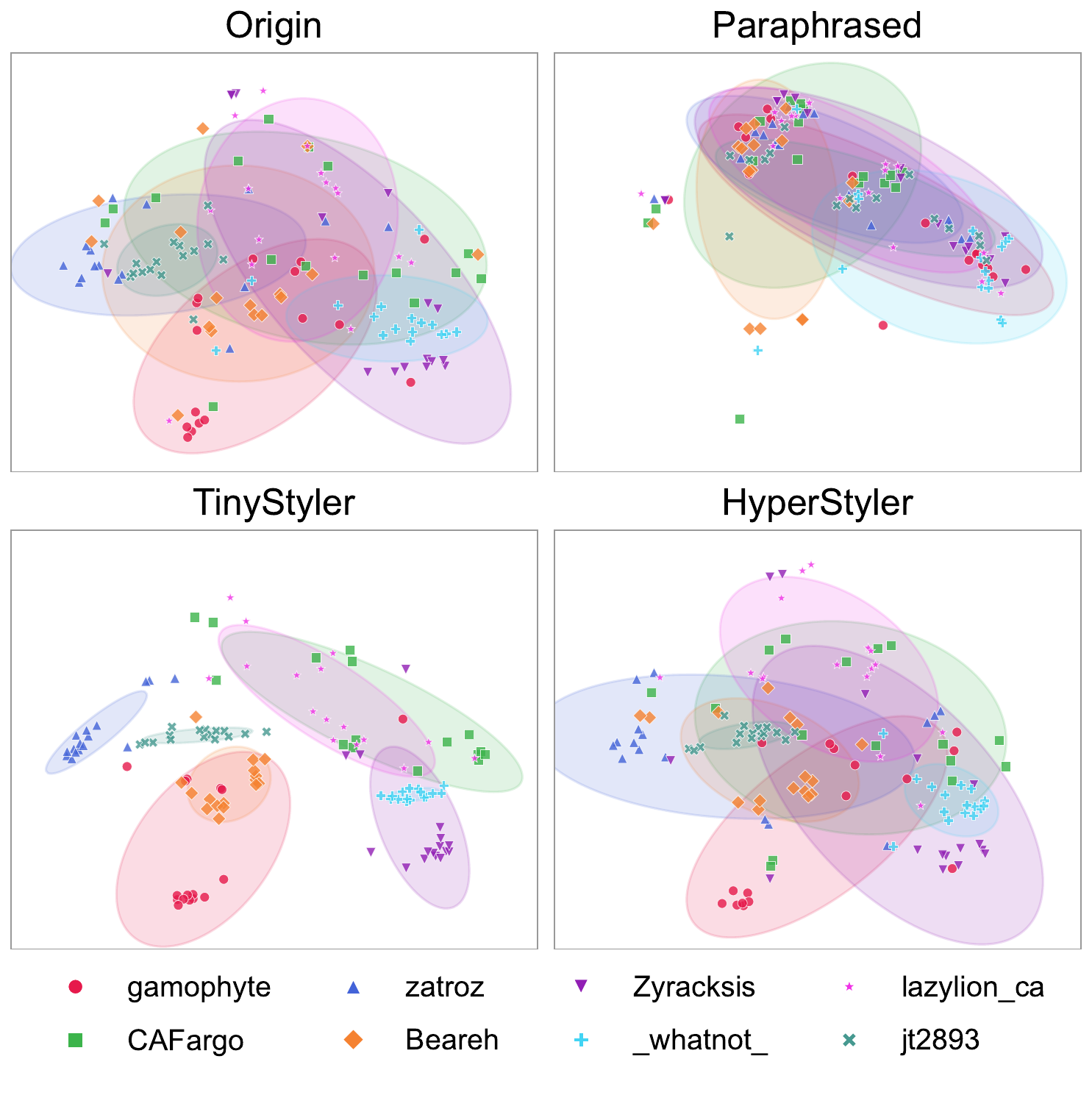}
    \caption{t-SNE visualization of the style embeddings 
for original texts, their paraphrases, and reconstructed outputs 
from TinyStyler and HyperStyler. Each ellipse indicates the approximate style 
distribution of an author.}
    \label{fig:tnse_visualization}

\end{figure}

\begin{figure}[!h]
    \centering
    \includegraphics[width=\linewidth]{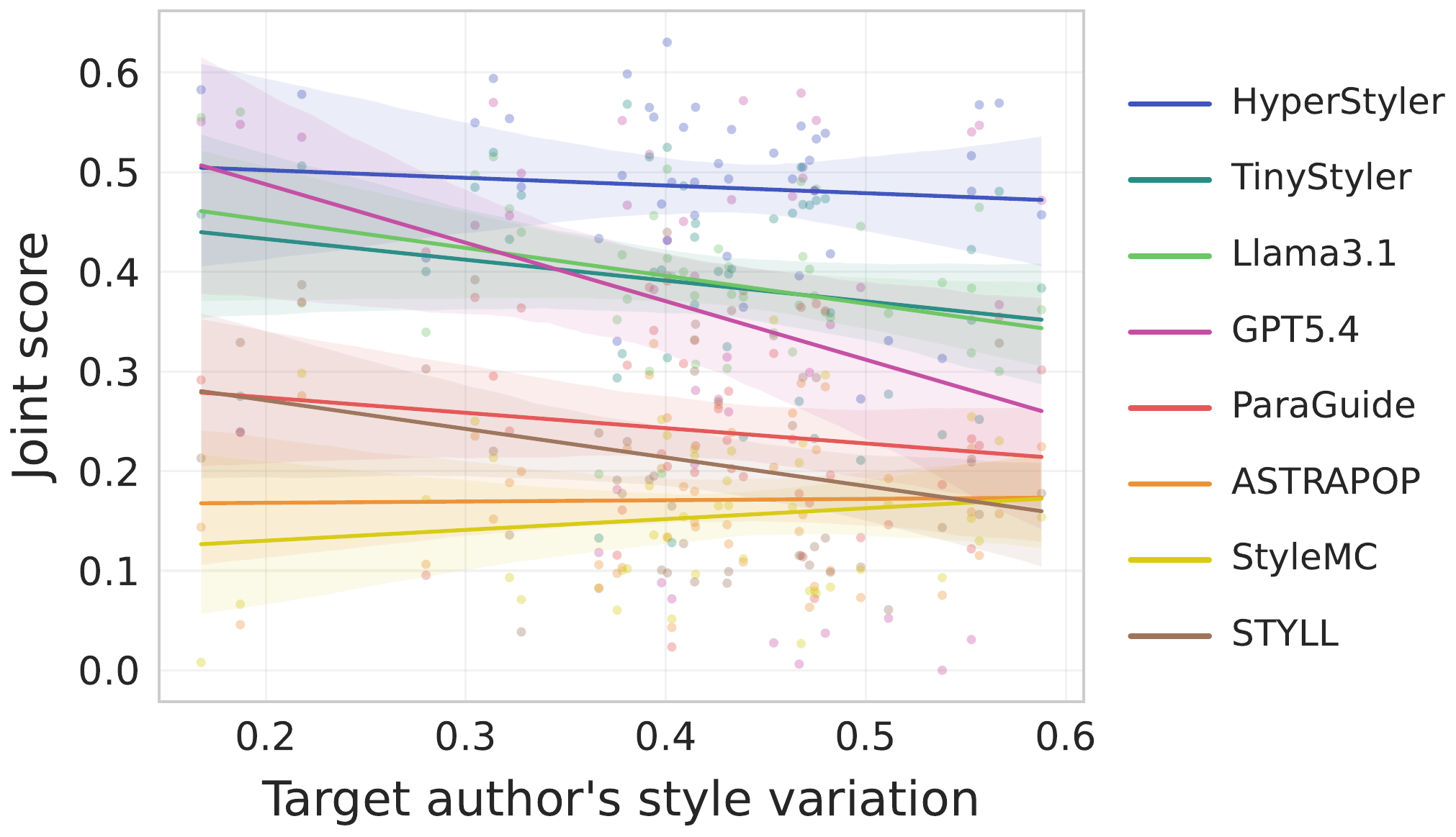}
    \caption{Trends of \textsc{Joint} score with respect to the target author’s style variation on Reddit. Lines indicate fitted linear trends with 95\% confidence bands.}
    \label{fig:trend_joint_by_style_variation}

\end{figure}

We examine whether HyperStyler navigates the style space in a context-dependent manner. Specifically, we paraphrase the original texts and have the model reconstruct them in their corresponding styles. As shown in Figure~\ref{fig:tnse_visualization}, TinyStyler, which is conditioned on a single static embedding, fails to faithfully reproduce the original stylistic distribution, whereas HyperStyler closely matches it. Furthermore, the predicted $z$ achieves a cosine similarity of 0.82 with the original style embedding and a Mean Reciprocal Rank (MRR) of 0.80, substantially outperforming mean pooling (cosine similarity:0.58, MRR: 0.21). These results demonstrate that the Stylo-navigator accurately identifies the context-appropriate style target. We also analyze performance under target-author style variation. Figure~\ref{fig:trend_joint_by_style_variation} shows that HyperStyler remains robust as variation increases, whereas the baselines degrade. This highlights that context-aware style selection enables HyperStyler to effectively handle high stylistic variation within an author's style.

\paragraph{Does style selection need to be explicit?} We compare the Stylo-navigator against two alternative strategies. (1) \textbf{Mean pooling}: We inject the mean-pooled reference embedding into the Stylo-hypernet. (2) \textbf{Implicit selection}: We provide all reference embeddings and perform layer-wise style selection implicitly via cross-attention. The results in Table~\ref{ablations} show that both alternatives exhibit substantially lower \textsc{Towards} scores compared to the Stylo-navigator, demonstrating that \textit{explicit} style selection conditioned on the source context is effective in improving style fidelity.

\begin{table}[!h] 
\centering
\resizebox{\columnwidth}{!}{%
\begin{tabular}{@{}llcccc@{}}
\toprule
\multicolumn{2}{l}{Model} & \textsc{Away} & \textsc{Towards} & \textsc{Sim} & \textsc{Joint} \\ \midrule
\multicolumn{2}{l}{HyperStyler} & 0.818 & \textbf{0.152} & 0.578 & \textbf{0.418} \\
 & \textit{w/o Stylo-navigator (Mean-pooling)} & 0.783 & 0.099 & 0.706 & 0.368 \\
 & \textit{w/o Stylo-navigator (Implicit selection)} & 0.778 & 0.114 & 0.671 & 0.384 \\
 & \textit{w/o Stylo-hypernet (Global)} & 0.990 & 0.006 & 0.165 & 0.016 \\
 & \textit{w/o Stylo-hypernet (Layer-wise)} & 0.791 & 0.121 & 0.630 & 0.394 \\
 & \textit{w/o adapter in FFN} & 0.800 & 0.134 & 0.608 & 0.409 \\
 & \textit{w/o prefix in CrossAttn} & 0.825 & 0.150 & 0.551 & 0.408 \\
 & \textit{w/ prefix in SelfAttn} & 0.969 & 0.016 & 0.461 & 0.076 \\
 & \textit{Underlying paraphraser} & 0.896 & 0.013 & 0.718 & 0.088 \\ 
 \bottomrule
\end{tabular}%
}
\caption{Ablation study on HyperStyler. We report the averaged value across three test splits of Reddit dataset.}
\label{ablations}
\vspace{-10pt}
\end{table}

\paragraph{Should style realization operate in parameter space?} We compare parameter modulation against two hidden-state injection strategies. (1) \textbf{Global}: We concatenate predicted $z$ to the encoder hidden states, providing the same style signal to all decoder layers. (2) \textbf{Layer-wise}: We generate layer-specific style embeddings and concatenate them to the encoder hidden states for each decoder layer. The global strategy fails to induce style changes, confirming that layer-wise style control is necessary. More importantly, while layer-wise injection shows some improvement, the \textsc{Towards}/\textsc{Sim} ratio of HyperStyler (0.263) is approximately 37\% higher than that of layer-wise injection (0.192), indicating that parameter modulation achieves higher style fidelity for the same semantic cost compared to hidden-state injection. This demonstrates the effectiveness of parameter modulation in realizing style while preserving content.

\paragraph{Which decoder components should be modulated?} We compare different combinations of modulation targets. Modulating only the cross-attention (\textit{w/o Adapter in FFN}) achieves high performance but falls short in style fidelity, while modulating only the FFN (\textit{w/o Prefix in CrossAttn}) improves style fidelity but degrades content preservation. Modulating the self-attention layers results in broken sentence structures, interfering with the decoder's generation process. These results suggest that jointly modulating FFN and cross-attention achieves the best balance between style fidelity and content preservation. 

\begin{figure}[h]
    \centering
    \includegraphics[width=\linewidth]{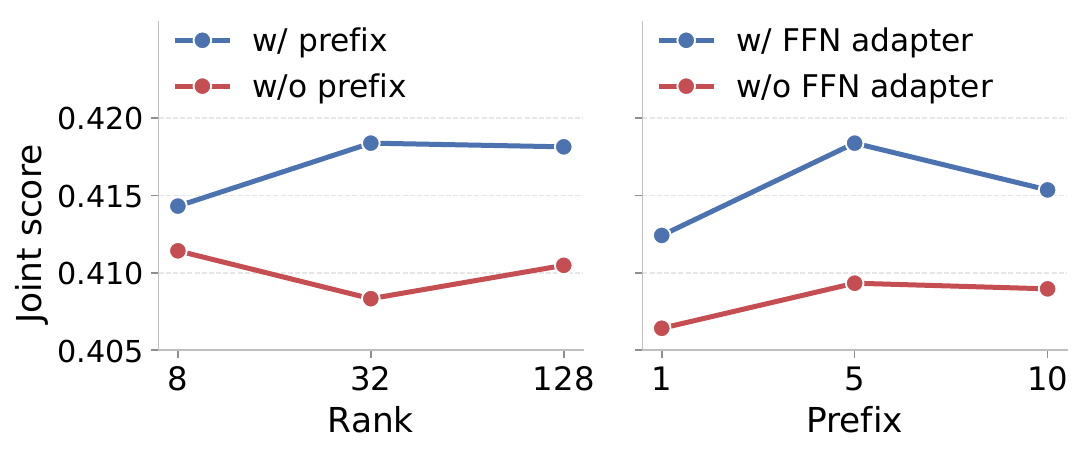}
    \caption{Effect of rank and prefix length with and without each modulation component, averaged across three test splits on the Reddit dataset.}
    \label{fig:ablation_hyperparams}
\vspace{-10pt}
\end{figure}

We further analyze the effects of varying rank and prefix length. As shown in Figure~\ref{fig:ablation_hyperparams}, increasing the rank yields only limited additional benefit, and a longer prefix does not necessarily yield further gains. Across all combinations of rank and prefix length, using both modulation components consistently outperforms variants that remove either the FFN adapter or the cross-attention prefix, further supporting the structural effectiveness of dual modulation.

\subsection{Human Evaluation}
We conduct a human evaluation of style transfer quality. Annotators evaluated two criteria: style fidelity (SF), where they selected whether the transferred text or the source text better matches the target author's style, and content similarity (CS), where they rated how well the transferred text preserves the meaning of the source text. More details are provided in Appendix \ref{human_eval_setup}. As shown in Table \ref{tab:human_eval_joint}, HyperStyler achieves the highest SF and G.mean scores. Its SF score is statistically significantly higher than those of GPT5.4 and ParaGuide. Meanwhile, its CS score is not significantly different from ParaGuide, which achieves the highest CS score. These results further validate that HyperStyler achieves high style fidelity while preserving content comparably to the baseline, demonstrating consistency across both automated metrics and human judgment.

\begin{table}[h]
\centering
\footnotesize
\resizebox{0.80\columnwidth}{!}{%
\begin{tabular}{@{}lccc@{}}
\toprule
Model & SF & CS & G.mean \\ \midrule
GPT5.4 & 0.47$^{\dagger}$ & 1.17 & 0.36 \\
Llama3.1 & \underline{0.59} & 1.07$^{\dagger\ddagger}$ & 0.41 \\
ParaGuide & 0.44$^{\dagger\ddagger}$ & \textbf{1.20} & 0.34 \\
TinyStyler & \underline{0.59} & 1.16 & \underline{0.44} \\
TinyStyler$_{\text{RERANK}}$ & 0.51 & 1.10 & 0.38 \\
HyperStyler & \textbf{0.61} & 1.15 & \textbf{0.46} \\
HyperStyler$_{\text{RERANK}}$ & 0.58 & \underline{1.19} & 0.43 \\ \bottomrule
\end{tabular}%
}
\caption{Human evaluation results. Bold and underlining indicate the best and second-best scores, respectively. ${\dagger}$ and ${\ddagger}$ denote significant differences from them, respectively. Significance is assessed at $p<0.05$ using McNemar's test for SF and paired t-tests for CS. G.mean is the geometric mean of SF and normalized CS.}
\label{tab:human_eval_joint}
\vspace{-8pt}
\end{table}

\subsection{Efficiency}
\paragraph{Parameter Efficiency.} We evaluate the parameter efficiency of our approach by testing a constrained configuration with a reduced rank and prefix length. As summarized in Table~\ref{ParameterEfficiency}, decreasing the number of parameters leads to a slight degradation in performance. Nevertheless, even with the rank and prefix length set to 1, our model still outperforms TinyStyler while adding only 2.4\% of the underlying model's parameters. This result demonstrates that HyperStyler maintains robust performance even under parameter constraints. 
\begin{table}[h]
\setlength{\tabcolsep}{2.5pt}
\centering
\resizebox{\columnwidth}{!}{%
\begin{tabular}{@{}lcccccccc@{}}
\toprule
Model & \textit{r} & \textit{p} & $\textsc{Away}$ & $\textsc{Towards}$ & $\textsc{Sim}$ & $\textsc{Joint}$ & \#Params & $\Delta$ \\ \midrule
TinyStyler & - & - & 0.860 & 0.122 & 0.626 & 0.399 & 783M & -  \\ \midrule
HyperStyler & 1 & 1 & 0.801 & 0.141 & 0.602 & 0.413 & 802M & $+2.4\%$  \\
 & 8 & 5 & 0.817 & 0.148 & 0.580 & 0.414 & 817M & $+4.3\%$  \\
 & 32 & 5 & 0.818 & 0.152 & 0.578 & \textbf{0.418} & 867M & $+10.7\%$  \\ \bottomrule
\end{tabular}%
}
\caption{Parameter efficiency analysis. \textit{r} and \textit{p} denote the adapter rank and prefix length, respectively. \#Params denotes the number of parameters and $\Delta$ represents the percentage increase relative to the T5-large model.}
\label{ParameterEfficiency}
\end{table}

\paragraph{Inference Time and Memory.} HyperStyler achieves approximately one second per inference on a single A100 GPU, with a modest overhead over TinyStyler (Table~\ref{table:inference_time}). Compared to open-source LLMs, HyperStyler is over 1.8$\times$ faster and uses less than one-eighth of the VRAM, and over 2.0$\times$ faster than API-based LLMs. Notably, even with reranking applied, HyperStyler remains faster than LLM-based baselines. These results demonstrate that high style transfer performance can be achieved without compromising computational efficiency, highlighting its suitability for time- and memory-constrained practical applications.

\section{Conclusion}
We introduce HyperStyler, a novel architecture for LAST grounded in the stylometric view that authorship style is not static but varies with context. HyperStyler explicitly decouples the task into two stages: context-aware style selection and stylistic realization. By explicitly selecting the most contextually appropriate style from a limited set of references and realizing it through parameter-space modulation, HyperStyler effectively addresses the mode averaging and style-content entanglement problems of existing methods. Extensive experiments demonstrate that HyperStyler consistently outperforms existing baselines and generalizes robustly across diverse domains, while maintaining superior performance even with only a 2.4\% increase in parameters. These results suggest that explicitly decoupling style selection and realization is a promising direction for achieving high-fidelity authorship style transfer in few-shot settings.

\section*{Limitations}
While HyperStyler achieves strong performance, we acknowledge several limitations stemming from the current LAST task setting. Our study primarily focuses on short-text transformation, typically consisting of one to three sentences, following the established protocols of the LAST benchmark \cite{patel2022low}. In practical applications, stylistic editing often extends beyond short texts to paragraph- or document-level inputs. However, paragraph-level authorship transfer requires dedicated solutions for defining and representing long-form stylistic elements as style conditions. Authorship style at this level encompasses compositional elements beyond sentence-level lexical and syntactic patterns, such as discourse structure, argument development, inter-sentence coherence, transitions, and narrative flow. Furthermore, extending authorship transfer to longer texts is hindered by the lack of reliable evaluation protocols for long-form style transfer, as current UAR-based evaluation metrics are primarily validated in short-text settings and may fail to capture cross-sentence stylistic coherence. These challenges highlight the need for future work on authorship transfer at the paragraph and document level, along with long-form evaluation protocols.

Another limitation is that our experiments are conducted on English corpora. Although HyperStyler is not inherently language-specific, extending LAST to multilingual or cross-lingual settings would require language-appropriate style representations capable of capturing content-independent stylistic signals across languages, as well as reliable evaluation protocols for assessing style fidelity and content preservation in multilingual settings. Addressing multilingual and cross-lingual authorship transfer remains an important direction for future work.

\section*{Ethics Considerations}
\paragraph{Potential Misuse and Impersonation:} LAST enables effective content personalization and stylistic imitation using only a few examples. However, this technique could be exploited by malicious actors for unauthorized impersonation. High-fidelity stylistic imitation, which HyperStyler achieves, poses a significant challenge to existing AI-generated text detection methods, suggesting that a new paradigm for authorship-aware detection is required to identify sophisticated synthetic texts. We advocate for the respectful use of stylistic imitation and emphasize that the responsibility for the generated content remains with the user.

\paragraph{Content Risks and Potential Bias:} Our training data includes datasets from online communities such as Reddit, which inherently contain offensive language, sexual content, or unethical sentiments. In this study, we did not apply explicit pre-filtering to the training data to preserve the raw stylistic features of the source domains. Consequently, the model may inadvertently generate unethical or biased outputs. We strongly advise that robust safety filters and post-processing mechanisms must accompany any real-world deployment of the model to prevent the dissemination of harmful content.

\section*{Acknowledgments}
This material is based upon work supported by the Air Force Office of Scientific Research under award number FA2386-23-1-4121, by the National Research Foundation of Korea (NRF) grant funded by the Korea government (MSIT) (RS-2024-00458720), and by the Institute of Information \& Communications Technology Planning \& Evaluation (IITP) grants funded by the Korean government (MSIT) (RS-2024-00439932, SW Starlab; No.RS-2020-II201336, Artificial Intelligence graduate school support (UNIST); No.RS-2021-II212068, Artificial Intelligence Innovation Hub; RS-2025-25442824, AI Star Fellowship Program (Ulsan National Institute of Science and Technology)). 
The authors used a generative AI tool for linguistic refinement and grammatical editing of the manuscript.


\bibliography{ref}

\clearpage
\appendix

\section{Data Description} \label{data_description}
\subsection{Reddit}
We use Million User Dataset (MUD) \cite{khan-etal-2021-deep}, a large-scale publicly available (Apache-2.0) user text dataset collected from the social media platform Reddit. The dataset comprises over 300 million Reddit posts produced by approximately one million users over the course of one year, and includes text-based user contributions in the form of comments. Evaluation is conducted on three predefined splits from \citet{patel2022low}.
\begin{itemize}
    \item \textbf{Diverse}: Source and target authors with posts on diverse topics across 13 or more different subreddits. \vspace{-8pt}
    \item \textbf{Random}: Random source and target authors.\vspace{-6pt}
    \item \textbf{Single}: All posts belong to a popular college football subreddit.
\end{itemize}

\subsection{Blog}
We use the Blog Authorship Corpus \cite{schler2006effects} collected from blogger.com, which consists of blog posts written by 19,320 individual bloggers. This dataset is freely available for non-commercial research purposes. This dataset is available at \url{https://www.kaggle.com/datasets/rtatman/blog-authorship-corpus/data}

\subsection{News} 
All-the-news dataset contains news articles collected from major U.S. and English-language news outlets. This dataset is available at \url{https://huggingface.co/datasets/rjac/all-the-news-2-1-Component-one}.

Since our objective is to analyze writing characteristics at the single-author level, we apply a series of filtering steps to ensure data quality. Specifically, we remove articles with missing author information, exclude articles attributed to organizations or non-individual entities, and discard articles with multiple authors. After filtering, only articles attributed to clearly identifiable individual authors are retained.

\begin{table}[h]
\centering
\small
\begin{tabular}{@{}ccccc@{}}
\toprule
       Dataset      & \#samples & \#authors & \#parallel pairs \\ \midrule
Reddit      & 7.5M  &  946K        &    200K   \\ 
Blog         & 177K   &   17K      &   40K       \\ 
News         & 538K &   53K        &     200K     \\
\bottomrule
\end{tabular}
\caption{Dataset statistics for training.}
\end{table}

\subsection{Stylistic Distance across Datasets}
Table \ref{tab:intra_inter_author_distance} reports intra-author style variation and inter-author distance for each dataset, providing supporting statistics for the performance differences observed across domains. Style variation is measured as the mean cosine distance of each author's style embeddings to their centroid, averaged across authors. Inter-author distance is measured as the mean pairwise cosine distance between authors in the UAR space.
\begin{table}[h]
\small
\centering
\begin{tabular}{@{}lcc@{}}
\toprule
Dataset & Style variation & Inter-author distance \\ \midrule
Reddit (Single)  & 0.407 & 0.310 \\
Reddit (Random)  & \textbf{0.414} & \textbf{0.384} \\
Reddit (Diverse) & 0.398 & 0.357 \\
Blog             & 0.316 & 0.322 \\
News             & 0.230 & 0.232 \\ \bottomrule
\end{tabular}%
\caption{Style variation and inter-author distance across datasets. Higher values reflect greater intra-author stylistic variability and greater inter-author separation.}
\label{tab:intra_inter_author_distance}
\vspace{-10pt}
\end{table}

\subsection{Cross-domain Inter-author Distance}
\begin{figure}[h]
    \centering
    \includegraphics[width=\linewidth]{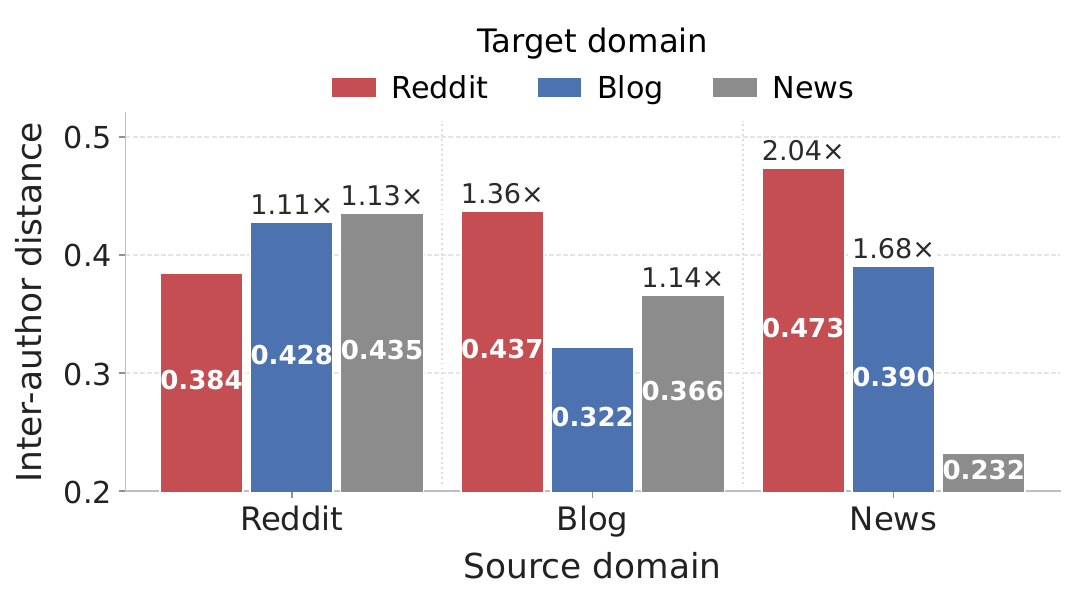}
    \caption{Cross-domain inter-author distances measured in the UAR space. Bars show inter-author distance, with annotations indicating the ratio relative to the in-domain distance.}
    \label{fig:cross_domain_inter_author}
\end{figure}

Figure \ref{fig:cross_domain_inter_author} shows the mean pairwise inter-author distances between source and target domain authors in the UAR space. Cross-domain distances are consistently larger than in-domain distances across all domain pairs. The largest differences relative to in-domain distance are observed for News to Reddit and Blog. Notably, since the \textsc{Towards} and \textsc{Away} metric is normalized by the source-target distance (Eq.~\ref{eq_towards}), larger inter-author distances naturally yield lower \textsc{Towards} values regardless of model performance. Consequently, \textsc{Towards} and \textsc{Joint} are not directly comparable across domain pairs with different inter-author distances, and should be interpreted in terms of relative differences between models within the same domain pair.

\section{Implementation Details} \label{implementation_details}
All training experiments are conducted on two NVIDIA A100 80GB GPUs. We use a batch size of 128 for all training stages. T5-large \cite{raffel2020exploring} is used as the base model for HyperStyler and all trainable baselines. For each method, we select the checkpoint with the lowest validation loss. At inference time, we use sampling with a temperature of 0.8 and top-$p$ to 1.0.

\subsection{HyperStyler}
For all attention layers in HyperStyler, we follow the T5 backbone by adopting a pre-norm structure with layer normalization and a multi-head decomposition \cite{vaswani2017attention} with $N_h=16$. We set $d=d_e$ for simplicity. For parameter efficiency, we share the projection matrices across all embedding tables.

\subsubsection{Selection of Style Embedding Space} \label{append:selection_style}
We adopt the STYLE embedder \cite{wegmann-etal-2022-author} to ensure a fair comparison with TinyStyler \cite{horvitz-etal-2024-tinystyler} and ParaGuide \cite{horvitz2024paraguide}, which also rely on it as the style conditioning signal. This embedder is trained via contrastive learning with negative samples from the same topic and domain, encouraging the model to capture subtle stylistic signals that distinguish authors within the same topic, yielding content-independent style representations. To empirically confirm this, we apply k-means clustering with Reddit dataset. Table \ref{tab:cluster_style_summary} exhibits the resulting clusters, which are clear and interpretable stylistic patterns, confirming that the STYLE embedder captures meaningful stylistic features beyond content.

\subsubsection{Selection of Activation Function for FFN Adapter}
We apply an activation function to the low-rank adapter used for FFN modulation (Eq.~\ref{eq_lora}). To examine whether an activation function is necessary and how the choice of activation function affects performance, we compare five variants: no activation, ReLU \cite{agarap2018deep}, ELU \cite{clevert2015fast}, GELU \cite{hendrycks2016gaussian}, and GeGLU \cite{shazeer2020glu}. As shown in Table~\ref{tab:activation_function}, the variant without an activation function exhibits relatively strong style transfer, but achieves the lowest overall performance due to lower semantic preservation. In contrast, variants using activation functions generally perform better than the no-activation variant. This suggests that an adapter with an activation function is more effective for balancing the trade-off between style fidelity and semantic preservation than a simple linear low-rank transformation.

Meanwhile, the performance differences among activation functions are relatively small. Therefore, we attribute the improvement primarily to the presence of an activation function rather than to any specific choice. Since GeGLU is used in the FFN of the underlying model (google/T5-v1.1-large) and achieves competitive performance, we adopt GeGLU in our model as well.

\begin{table}[h]
\centering
\small
\resizebox{0.86\columnwidth}{!}{%
\begin{tabular}{@{}lcccc@{}}
\toprule
Activation & \textsc{Away} & \textsc{Towards} & \textsc{Sim} & \textsc{Joint} \\ \midrule
w/o ACT & 0.881 & 0.165 & 0.454 & 0.386 \\
ReLU & 0.819 & 0.148 & 0.565 & 0.415 \\
ELU & 0.821 & 0.150 & 0.565 & 0.415 \\
GeLU & 0.827 & 0.151 & 0.560 & 0.411 \\
GeGLU & 0.818 & 0.152 & 0.578 & 0.418 \\ 
\bottomrule
\end{tabular}%
}
\caption{Effect of activation function choice in the FFN adapter. 'w/o ACT' denotes the setting without an activation function (no activation).}
\label{tab:activation_function}
\vspace{-5pt}
\end{table}

\begin{table}[h]
\centering
\resizebox{\columnwidth}{!}{%
\begin{tabular}{@{}l|ccc@{}}
\toprule
Configuration & Stage 1 & Stage 2 & Stage 3 \\ \midrule
Learning rate & $5e^{-5}$ & $1e^{-4}$ & $1e^{-4}$ \\
Batch size & 128 & 128 & 128 \\
Optimizer & AdamW & AdamW & AdamW \\
Weight decay & 0.01 & 0.01 & 0.01 \\ 
Scheduler & Constant & Cosine & Constant \\
Warm-up steps & 2000 & 2000 & 5\% of max steps \\
Max steps / epochs & 200K steps & 100K steps & 3 epochs \\ \bottomrule
\end{tabular}%
}
\caption{Training setup for HyperStyler.}
\label{tab:HyperStyler_setup}
\vspace{-5pt}
\end{table}

\subsection{TinyStyler}
We follow the original paper's configuration and use the provided training code \cite{horvitz-etal-2024-tinystyler}. Only for the Reddit dataset, we use the publicly available checkpoint rather than training from scratch, as we found that training from scratch in our environment yielded lower performance than originally reported.

\begin{table}[h]
\centering
\fontsize{9.5pt}{11.5pt}\selectfont
\begin{tabular}{@{}l|c@{}}
\toprule
Configuration & Value \\ \midrule
Pretrained Ckpt & google/t5-v1\_1-large \\
Learning rate & $1e^{-5}$ \\
Batch size & 128 \\
Optimizer & Adam \\
Weight decay & 0.01 \\
Schedule & Constant \\
Warm-up Steps & 2000 \\
Total Steps & 150K \\ \bottomrule
\end{tabular}%
\caption{Hyperparameters of TinyStyler.}
\label{tab:my-table}
\end{table}

\subsection{StyleMC}
Since the official source code for StyleMC \cite{khan2023learning} is not publicly available, we implemented the method by strictly adhering to the descriptions provided in the original paper. While we made every effort to ensure a faithful reproduction, minor discrepancies may exist compared to the original implementation due to unspecified details of the algorithm or hyperparameters. For a fair comparison, we modified the baseline STYLEMC by replacing its original UAR-based implementation with the STYLE embedder \cite{wegmann-etal-2022-author}. An author embedding was then calculated via mean pooling, aligning it with the evaluation protocol used for other models.

\begin{table}[h]
\centering
\fontsize{9.5pt}{11.5pt}\selectfont
\begin{tabular}{@{}l|c@{}}
\toprule
Configuration & Value \\ \midrule
Learning rate & $1e^{-5}$ \\
Batch size & 128 \\
Optimizer & AdamW \\
Weight decay & 0.01 \\
Future discriminator Ckpt & facebook/opt-1.3b \\
Proposal generator Ckpt & google/t5-v1\_1-large \\
Number of steps & 80$\times$Sequence length \\
$\alpha_{\text{fluency}}$ & 0.005 \\
$\alpha_{\text{style}}$ & 1.0 \\
$\alpha_{\text{semantic}}$ & 1.0 \\
$\alpha_{\text{edit}}$ & 0.01 \\ \bottomrule
\end{tabular}%
\caption{Hyperparameters of StyleMC.}
\label{tab:my-table}
\end{table}

\subsection{ASTRAPOP}
We conduct experiments based on the ASTRAPOP framework \cite{liu2024authorship} with CPO \cite{xu2024contrastive}, its best-performing variant, while modifying several components to better align it with our experimental setting. Originally, ASTRAPOP uses LLaMA-2-7B, a decoder-only model, as its backbone architecture. To ensure a fair evaluation, we replace the backbone with T5-Large. This architectural change requires decisions on how the source and reference texts are arranged in the encoder input. We also explore the JOINT reward formulation, following the filtering criterion used in TinyStyler, to examine whether it provides a more effective training signal than the original reward. To identify the configuration that performs best under our setting, we evaluate four variants combining input order and reward function:
\begin{itemize}
\item \textbf{ASTRAPOP} follows the original input order and reward formulation.
\item \textbf{ASTRAPOP${_\text{\textsc{JOINT}}}$} adopts the JOINT reward formulation while preserving the original input order.
\item \textbf{ASTRAPOP${_\text{reverse}}$} places the \texttt{[src]} token at the beginning of the input sequence, while the original reward formulation remains unchanged.
\item \textbf{ASTRAPOP$_{\text{reverse, \textsc{Joint}}}$} combines both the reversed input order and the JOINT reward formulation.
\end{itemize}
According to Table~\ref{table12}, no configuration performs consistently best across datasets. We therefore report the two variants that preserve the original input order in Table~\ref{table1}.

In addition, to examine whether the backbone replacement puts ASTRAPOP at a disadvantage, we train ASTRAPOP with its original LLaMA-2-7B backbone and compare it with HyperStyler. As shown in Table~\ref{ASTRAPOP_LLaMA}, despite a more than 9$\times$ difference in model size, HyperStyler consistently achieves higher \textsc{Joint} scores across three datasets.

\begin{table}[h]
\centering
\fontsize{9.5pt}{11.5pt}\selectfont
\begin{tabular}{@{}l|cc@{}}
\toprule
Configuration & SFT & CPO \\ \midrule
learning rate & $5e^{-5}$ & $1e^{-5}$ \\
batch size & 128 & 128 \\
Optimizer & Adam & Adam \\
\# epochs & 20 & 20 \\
Max steps & 100K & 100K \\
$\beta$ & -- & 0.1 \\
top $p$ & -- & 1.0 \\
temperature & -- & 0.8 \\
length penalty $\alpha$ & -- & 0.5 \\
Context Size & 512 & 512 \\
Output Size & 80 & 80 \\ \bottomrule
\end{tabular}%

\caption{Hyperparameters of ASTRAPOP.}
\label{tab:my-table}
\end{table}

\subsection{Paraguide}
We followed the experimental setup used in the original paper \cite{horvitz2024paraguide}.

\begin{table}[h]
\centering
\fontsize{9.5pt}{11.5pt}\selectfont
\begin{tabular}{@{}l|c@{}}
\toprule
Configuration & Value \\ \midrule
Pretrained Ckpt & xhan77/ssdklm \\
Learning rate & $5 \times 10^{-6}$ \\
Batch size & 128 \\
Optimizer & AdamW \\
Weight decay & 0.01 \\
Schedule & Constant \\
Warm-up Steps & 2000 \\
Total Steps & 150K \\
Diffusion Steps & 200 \\
Context Size & 80 \\
Output Size & 80 \\ \bottomrule
\end{tabular}%
\caption{Hyperparameters of Paraguide.}
\label{tab:my-table}
\end{table}

\subsection{In-context Learning Methods}
For GPT-based models and Llama-3.1, we use the prompt from \citet{horvitz-etal-2024-tinystyler}, with default API settings for GPT-4-turbo (gpt-4-turbo-2024-04-09), GPT-5-mini (gpt-5-mini-2025-08-07), and GPT-5.4 (gpt-5.4-2026-03-05, medium). STYLL follows the original experimental setup \cite{patel2022low} using Qwen2.5-7B.

\section{Evaluation Details} 
\subsection{Metric Formula Definition} \label{Metric_Formula} We adopt the evaluation metrics proposed by \citet{patel2022low} for LAST. For any author $a$, let $P_a$ denote their set of 16 posts, and $P_{s \rightarrow t}$ denote the set of posts written by source author $s$ and style-transferred to target author $t$. Let $\vec{R}(P)$ denote a single UAR embedding produced over a set of posts $P$. Finally, we define $S(\mathbf{u}, \mathbf{v})$ scaled to the range $[0,1]$, given by $S(\vec{u}, \vec{v}) = \frac{\mathrm{sim}(\vec{u},\vec{v}) + 1}{2}$. We further define its complement as $S_c(\vec{u},\vec{v}) = 1 - S(\vec{u},\vec{v})$.

\paragraph{\textsc{Away}} measures how far a style-transferred text departs from the source author’s style:
\begin{equation}
\frac{
\min\Big(
S_c\big(\vec{R}(P_{s\to t}), \vec{R}(P_s)\big),\,
S_c\big(\vec{R}(P_t), \vec{R}(P_s)\big)
\Big)
}{
S_c\big(\vec{R}(P_t), \vec{R}(P_s)\big)
}
\label{eq_away}
\end{equation}

\paragraph{\textsc{Towards}} measures how far a style-transferred text moves toward the target author’s style:
\begin{equation}
\frac{
\max\Big(
S\big(\vec{R}(P_{s\to t}), \vec{R}(P_t)\big)
-
S\big(\vec{R}(P_s), \vec{R}(P_t)\big),
0
\Big)
}{
S_c\big(\vec{R}(P_s), \vec{R}(P_t)\big)
}
\label{eq_towards}
\end{equation}

\paragraph{\textsc{Sim}} measures how well the transferred text preserves the meaning of the source text. The average Mutual Implication Score \cite{babakov-etal-2022-large} between two sets of posts authored by $a$ and $b$ is denoted as $\mathrm{MIS}(P_a, P_b)$:
\begin{equation}
\frac{
\max\Big(
\mathrm{MIS}(P_{s\to t}, P_s)
-
\mathrm{MIS}(P_t, P_s),
0
\Big)
}{
1 - \mathrm{MIS}(P_t, P_s)
}
\end{equation}

\subsection{Details on Human Evaluation} \label{human_eval_setup}
We recruit annotators from Amazon Mechanical Turk, restricting participation to workers from English-speaking countries (i.e., US, UK, Canada, Australia) with a 95\% or higher approval rating. As the evaluation of style transfer is known to be difficult for humans \cite{krishna-etal-2020-reformulating, patel2022low, hallinan-etal-2023-steer, liu2024authorship}, we introduce a qualification test to ensure a minimum level of annotation quality. The test consists of three items. In each item, annotators are shown five reference texts from each of two randomly selected authors and asked to identify which author wrote a held-out target text. Only annotators who correctly answer all three items are admitted to the main evaluation.

For each baseline category, we select the best-performing model for the main evaluation. The evaluation is conducted on the same 100 source-target author pairs sampled from the Reddit test set, with three annotators assigned to each model output. We exclude examples whose source texts or target-author references contain violent, sexually explicit, or profane content to minimize annotator exposure to potentially harmful or offensive material. Annotators are also informed before the task that they may encounter potentially harmful or offensive content, and only those who agree to proceed participate in the evaluation. We pay 60 cents per annotated pair, corresponding to an estimated hourly rate based on the average completion time.

For style fidelity, annotators are shown eight reference texts from the target author, along with an anonymized source text and a transferred text in randomized order. They are then asked to select which text is more likely to have been written by the target author. The final label is determined by majority vote (Krippendorff's $\alpha= 0.10$), with a score of 1 assigned when the transferred text is selected and 0 otherwise. For content similarity, annotators are shown the source text and the transferred text and asked to rate their semantic similarity on a 3-point Likert scale: 0 indicates Not Similar, 1 indicates Somewhat Similar, and 2 indicates Similar. The average score across annotators is used as the final content similarity score. Detailed instructions are shown in Tables \ref{tab:sf_instruction} and \ref{tab:cs_instruction}. Our content similarity question and rating scale are adapted from \citet{liu2024authorship}. We use McNemar's test \cite{mcnemar1947note} for style fidelity and paired t-test for content similarity to verify whether performance differences between models are statistically significant, with $p<0.05$.

\begin{table}[th]
\centering
\fontsize{10pt}{11.5pt}\selectfont
\begin{tabular}{p{0.95\linewidth}}
\toprule
\textbf{Instruction}\\
Read the reference author's writing samples, then decide which of 
the two texts is more likely written by that author based on \textbf{writing 
style} (sentence structure, word choice, tone --- not topic).\\
\midrule
\textbf{Reference Author's Writing Samples}\\[4pt]
\textit{[Writing samples are shown here]}\\
\midrule
\textbf{Text A:} \textit{[Text A is shown here]}\\[4pt]
\textbf{Text B:} \textit{[Text B is shown here]}\\
\midrule
\textit{Which text is more likely written by the Reference Author, 
based on writing style?}\\[4pt]
$\square$ \textbf{Text A}\\[4pt]
$\square$ \textbf{Text B}\\
\bottomrule
\end{tabular}
\caption{Instruction for style fidelity evaluation.}
\label{tab:sf_instruction}
\end{table}

\begin{table}[th]
\centering
\fontsize{10pt}{11.5pt}\selectfont
\begin{tabular}{p{0.95\linewidth}}
\toprule
\textbf{Instruction}\\
Read both texts and judge how similar they are. Focus on the \textbf{core content} and \textbf{key information} conveyed, not the writing style.\\
\midrule
\textbf{Text A:} \textit{[Text A is shown here]}\\[4pt]
\textbf{Text B:} \textit{[Text B is shown here]}\\
\midrule
\textit{How similar are the two texts?}\\[4pt]
\textbf{0 — Not Similar}\\
Only small portions (less than 50\%) of the passages are the same.\\[4pt]
\textbf{1 — Somewhat Similar}\\
Large portions (50--75\%) of the passages are the same, but there are 
significant sections that differ or are present in only one passage.\\[4pt]
\textbf{2 — Similar}\\
Most of the content (75\% or more) of the two passages is the same.\\
\bottomrule
\end{tabular}
\caption{Instruction for content similarity evaluation.}
\label{tab:cs_instruction}
\end{table}

\section{Additional Experimental Results}\label{additional_experimental_results}

\subsection{Analysis on the Number of References}
\begin{figure}[th]
    \centering
    \includegraphics[width=\linewidth]{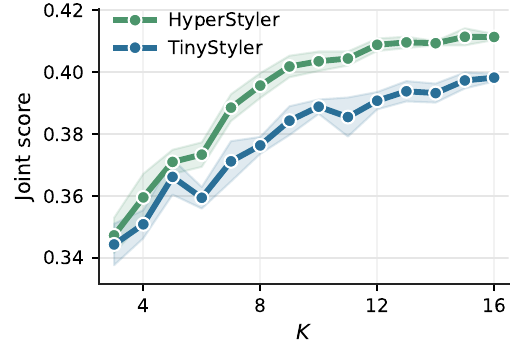}
    \caption{Trend of $\textsc{Joint}$ score across $K$ on the Reddit Random split, with shaded bands indicating standard error of the mean over five random samplings.}
    \label{fig:k_analysis}
\end{figure}

We analyze how performance varies with the number of references $K$. As shown in Figure~\ref{fig:k_analysis}, HyperStyler shows a consistent and substantial performance advantage over TinyStyler from $K=6$. Notably, HyperStyler with only $K=9$ references surpasses TinyStyler’s best performance at $K=16$. This suggests that HyperStyler uses the available references more effectively through context-aware style selection. These results indicate that the key factor is not simply the number of references, but how the model selects and uses stylistic evidence relevant to the source context.

\begin{table}[!h]
\centering
\setlength{\tabcolsep}{2pt}
\fontsize{10pt}{11.5pt}\selectfont
\begin{tabular}{@{}llcccc@{}}
\toprule
\multicolumn{2}{l}{Method} & \textbf{} & \textbf{} & Time(s) & VRAM(GiB) \\ \midrule
\multicolumn{2}{l}{\textit{In-context learning}} &  &  &  &  \\
 & STYLL (Qwen2.5-7B) &  &  & 14.85 & 32.3 \\
 & GPT-4 Turbo &  &  & 2.01 & - \\
 & GPT-5 Mini &  &  & 5.14 & - \\
 & GPT-5.4 &  &  & 8.50 & - \\
 & Llama-3.1-8B-Instruct &  &  & 1.83 & 30.8 \\ \midrule
\multicolumn{2}{l}{\textit{Inference-time control}} &  &  &  &  \\
 & ParaGuide$_{\lambda=200}$ &  &  & 20.93 & 3.40 \\
 & ParaGuide$_{\lambda=2500}$ &  &  & 20.86 & 3.40 \\
 & StyleMC &  &  & 49.51 & 9.29 \\ \midrule
\multicolumn{2}{l}{\textit{Unsupervised alignment}} &  &  &  &  \\
 & ASTRAPOP &  &  & 1.72 & 3.55 \\
 & TinyStyler &  &  & 0.82 & 3.17 \\
 & TinyStyler$_{\text{RERANK(5)}}$ &  &  & 1.15 & 5.05 \\ \midrule
\multicolumn{2}{l}{\textit{Proposed method}} & \textit{r} & \textit{p} &  &  \\
 & HyperStyler & 1 & 1 & 0.94 & 3.80 \\
 & HyperStyler$_{\text{RERANK(5)}}$ & 1 & 1 & 1.26 & 5.08 \\
 & HyperStyler & 8 & 5 & 1.01 & 3.80 \\
 & HyperStyler$_{\text{RERANK(5)}}$ & 8 & 5 & 1.45 & 5.13 \\
 & HyperStyler & 32 & 5 & 1.03 & 3.85 \\
 & HyperStyler$_{\text{RERANK(5)}}$ & 32 & 5 & 1.45 & 5.26 \\ \bottomrule
\end{tabular}
\caption{Inference cost. \textit{r} and \textit{p} denote the adapter rank and prefix length, respectively. Time denotes the average inference time over 300 instances in the Reddit test dataset, and VRAM denotes peak memory usage.}
\label{table:inference_time}
\end{table}

\subsection{Computational Cost Analysis}
Table \ref{table:inference_time} reports inference latency and memory usage on the Reddit test set. Inference time is averaged over 300 instances, and VRAM is measured as peak FP32 memory usage for locally hosted models on a single NVIDIA A100 GPU. For API-based LLMs, we report wall-clock latency only, as server-side memory usage is not accessible. For reranking variants, the reported time includes both candidate generation and reranking.

\begin{table}[!h]
\centering
\fontsize{10pt}{11.5pt}\selectfont
\begin{tabular}{@{}llccc@{}}
\toprule
\multicolumn{2}{l}{Training stage} & GPUs & Time \\ \midrule
\multicolumn{4}{l}{\textit{HyperStyler}} \\
 & Stage1 & A100 80G x2 & 28h \\
 & Stage2 & A100 80G x2 & 16h \\
 & Stage3 & A100 80G x2 & 0.5h \\ \midrule
\multicolumn{4}{l}{\textit{TinyStyler}} \\
 & Training & A100 80G x2 & 42h \\
 & Self-distillation & A100 80G x2 & 4h \\ \midrule
\multicolumn{4}{l}{\textit{ASTRAPOP}} \\
 & SFT & A100 80G x2 & 45h \\
 & CPO & A100 80G x2 & 24h \\ \midrule
\multicolumn{4}{l}{\textit{Paraguide}} \\
 & Finetuning & A100 80G x2 & 14h \\ \midrule
\multicolumn{4}{l}{\textit{StyleMC}} \\
 & Future regressor & A100 80G x2 & 18h \\ \bottomrule
\end{tabular}
\caption{Training cost on the Reddit dataset under each method's training configuration.}
\vspace{-5pt}
\label{tab:training-cost}

\end{table}

Table \ref{tab:training-cost} reports training times on the Reddit dataset. These times characterize the practical computational cost under our experimental setup rather than provide a strictly controlled comparison of training efficiency, since methods differ in training objectives, optimization hyperparameters, and training schedules. We omit in-context learning baselines because they do not require task-specific training. HyperStyler takes 28h, 16h, and 0.5h for its three stages, respectively. Its final self-distillation stage uses a filtered set of roughly 40K instances, similar to TinyStyler's, but requires less training time under our configuration.

\subsection{Qualitative Analysis}
\paragraph{Target-dependent style transfer.} Figure \ref{fig:target-dependent-style-transfer} visualizes t-SNE projections of style embeddings for the same source texts transferred from source author A to two target authors B and C. HyperStyler's A$\rightarrow$B and A$\rightarrow$C outputs occupy distinct stylistic regions and are more closely aligned with the corresponding target author's style variation. In contrast, TinyStyler and ParaGuide, which rely on static author embeddings, tend to concentrate in a particular stylistic region rather than aligning with the target-author references. Moreover, baselines that receive all reference texts as input also show weaker target-wise separation or weaker alignment with the corresponding target author's style. These results suggest that HyperStyler's explicit style navigation enables target-dependent style transfer, producing outputs that reflect the distinct stylistic characteristics of each target author.

Table \ref{tab:qualitative_target_dependent} presents examples where each source text is rewritten using two different target-author reference sets. Source 1 is an argumentative reply. For Target author A, the output opens with a question that reflects the question-based style observed in the references (\textit{What is your point?}), then restates the original advice in a more explicit form. For Target author B, the output stays relatively close to the source wording while reflecting the ellipsis usage observed in the references (\textit{personally....}). Source 2 is a reassurance-oriented comment. Target author A makes the response warmer and more supportive (\textit{I'm glad you found it}, \textit{:D}), whereas Target author B keeps it concise and neutral without an emotive marker. Source 3 is an event recap. For Target author A, the output adopts a more casual, punctuation-heavy recap style (\textit{yep..}, \textit{!!}). For Target author B, the output becomes a more straightforward recap, consistent with the more explanatory phrasing observed in the references. These examples suggest that HyperStyler can realize target-author-specific style cues differently while preserving the source context.

\begin{table}[h]
\fontsize{10pt}{11.5pt}\selectfont
\centering
\begin{tabular}{p{0.95\linewidth}}
\toprule
\textbf{Case 1: content omission under compressive style} \\
\midrule
\textbf{Source:} Yeah, the difference between the two ranks is pretty minimal. I mentioned a few weeks back that I didn't think we'd get a true gauge on Michigan State until the Notre Dame game ... I watched that entire game ... \\
\midrule
\textbf{Output:} i didn't think we'd get a true gauge on Michigan State until the Notre Dame game, but i definitely stand by that after the Furman game lmao \\
\midrule
\textbf{Case 2: over-generation under elaborative style} \\
\midrule
\textbf{Source:} Which is why I don't respect early season rankings ... AT ALL \\
\midrule
\textbf{Output:} I'm not a big fan of early season rankings, but I'm glad someone is enjoying the game. AJ Green is a good player \\
\bottomrule
\end{tabular}
\caption{Representative failure cases.}
\label{tab:failure}
\end{table}

\paragraph{Failure mode.}
We identified two representative failure modes, shown in Table~\ref{tab:failure}, that arise when the information density of the source text is misaligned with the stylistic signals provided by the reference set. The first is content omission under compressive target style, which occurs when the source text is long and information-dense while the reference set reflects a short, reaction-oriented conversational style. In such cases, HyperStyler tends to compress the source text, retaining the main stance but omitting secondary propositions and supporting details. The second is over-generation under elaborative target style, which occurs when the source text is short and self-contained while the reference set reflects a more expressive and elaborative style. In such cases, the model tends to expand the output to realize the target style, introducing content that is not grounded in the source text and potentially leading to semantic drift. These cases suggest that authorship style transfer becomes particularly challenging when the amount of information that must be preserved from the source conflicts with the degree of compression or elaboration implied by the target reference style.

\subsection{Analysis on Style-dependent and Layer-wise Modulation}

\begin{figure}[!ht]
    \centering
    \includegraphics[width=\linewidth]{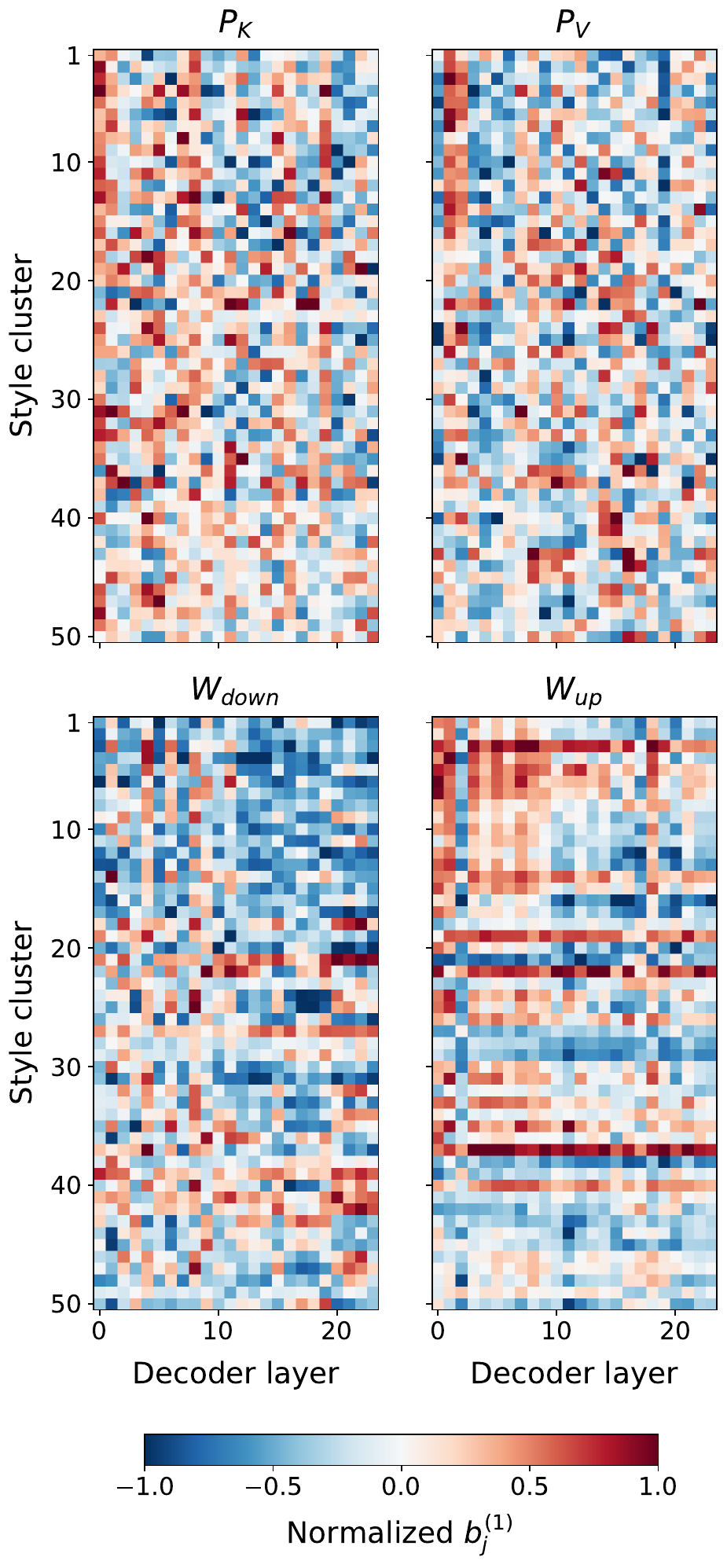}
    \caption{Compatibility scores $b_j^{(1)}$ for the first head across 50 style clusters and decoder layers. Top: cross-attention prefixes ($P_K$, $P_V$, first prefix position). Bottom: FFN low-rank projections ($W_{down}$, $W_{up}$).}
    \label{fig:layerwise_modulation_heatmap}
\end{figure}

\begin{figure*}[!ht]
    \centering
    \includegraphics[width=\textwidth]{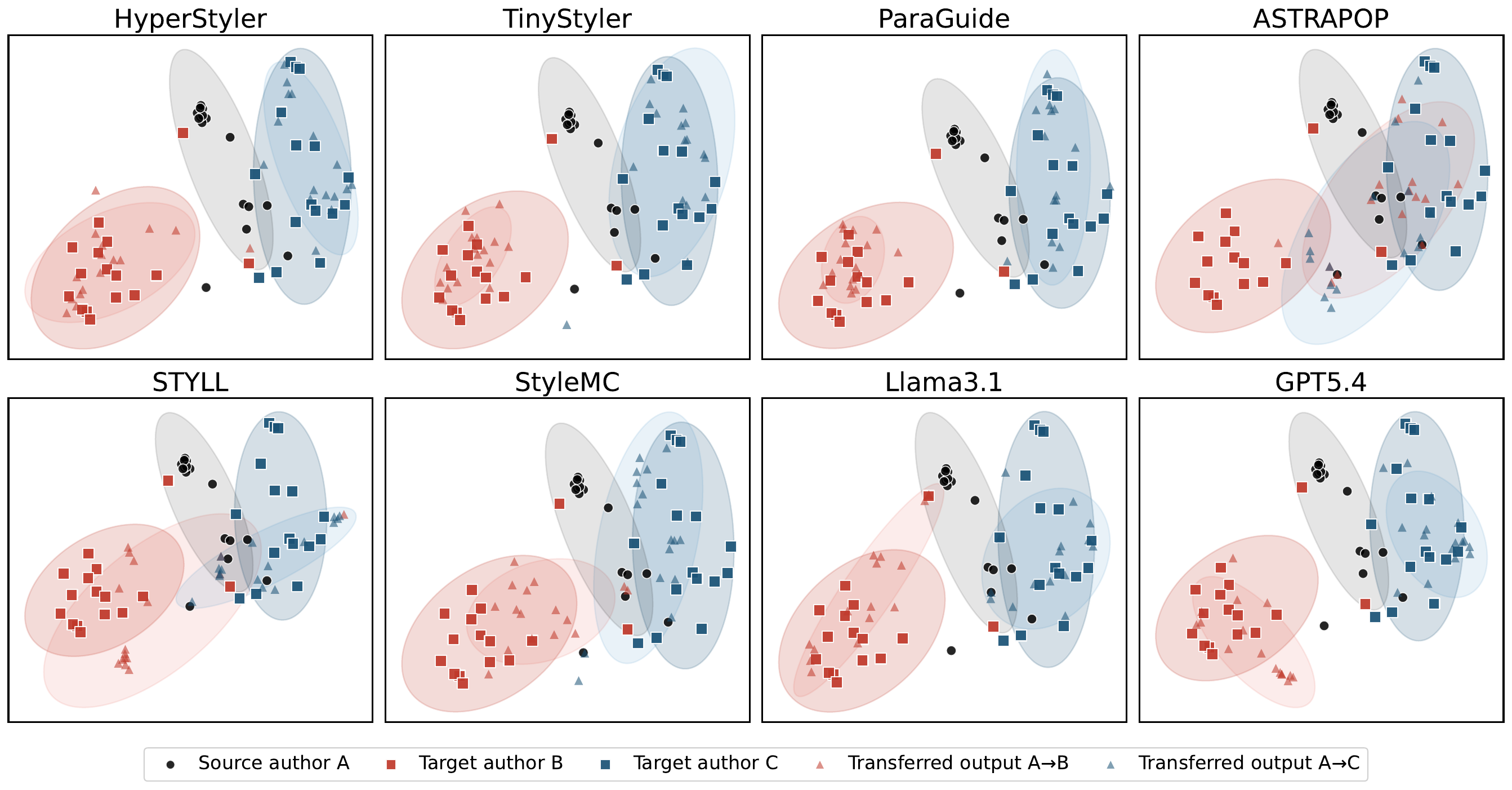}
    \caption{t-SNE visualization of style-transferred outputs across methods. Each panel shows how the stylistic distribution of transferred outputs shifts from the source author texts toward two target authors' styles (Target author B and Target author C), where ellipses indicate the approximate stylistic distribution boundary of each group.}
    \label{fig:target-dependent-style-transfer}
\end{figure*}

Stylo-hypernet modulates each decoder layer via a bilinear interaction between the style coordinate and learnable layer embeddings (Eq.~\ref{eq4}), yielding compatibility scores for each modulation target. We examine whether these scores vary across styles and layers. Using the k-means clustering results in Section~\ref{append:selection_style}, we select the 50 sentences closest to each cluster centroid. Each sentence is fed into the Stylo-hypernet to obtain its compatibility scores $b_j^{(h)}$, which are averaged within each cluster. The resulting scores are normalized per layer by the maximum absolute value across clusters. As shown in Figure~\ref{fig:layerwise_modulation_heatmap}, within a given layer, scores differ across style clusters. Within a given cluster, the scores also change from layer to layer. These results confirm that Stylo-hypernet assigns distinct scores across styles and layers, as intended by its design.

\section{Licenses and Use of Artifacts}
Table~\ref{tab:artifacts} lists artifacts used in this work, including models for training and evaluation and software libraries, with their licenses and links to the sources. Our use of the artifacts is consistent with their licenses and intended use. In particular, artifacts released under permissive licenses are used for research in accordance with their terms, and the artifact licensed under CC BY-NC-SA 4.0 is used only in a non-commercial research context. We do not redistribute any artifact in a manner inconsistent with its original license or access conditions, and any outputs or derived materials from this work are intended only for research use.

\clearpage

\begin{table*}[!t]
\centering
\scriptsize
\setlength{\tabcolsep}{4pt}
\renewcommand{\arraystretch}{1.12}
\begin{tabular}{p{0.48\textwidth}p{0.47\textwidth}}
\toprule
\textbf{Target author references} & \textbf{HyperStyler output} \\
\midrule[0.8pt]

\addlinespace[3pt]
\multicolumn{2}{p{0.96\textwidth}}{\textbf{Source 1:} Why do you need to address him personally at all? Attack the argument, not the person.} \\
\addlinespace[2pt]
\midrule
\textbf{Target author A:}
\textit{Why not? Im not a Nasus main but Rylai procs on his ult and E.};
\textit{Uhm hello what year is this? Even if hes in a straight relationship, he shouldnt do that.};
\textit{Can u explain the reference?};
\textit{DAE also think masc is the most opressed!!!?};
\textit{TIL the internet is one person}
&
What is your point? It is an argument, not a person. Attack the argument not the person \\
\midrule
\textbf{Target author B:}
\textit{Big difference between being a fanboi and simply not falling inline with the circlejerk....};
\textit{Don't worry /r/cars tells them everyday....};
\textit{For \$5,000 all you should expect is a pile of rusty parts...};
\textit{Wouldn't that be the Skrillex cut?};
\textit{This thread officially proves we are no better than the YouTube comments section}
&
Why are you even addressing him personally.... Attack the argument not the person. \\
\midrule[0.8pt]

\addlinespace[3pt]
\multicolumn{2}{p{0.96\textwidth}}{\textbf{Source 2:} No prob! Easy to make the mistake. Not like Reddit is graded or anything. :)} \\
\addlinespace[2pt]
\midrule
\textbf{Target author A:}
\textit{I guess I'm building. Thank you for your help! I really appreciate it!};
\textit{You guys have been really polite, I'm really surprised! Thank you for all your help so far!};
\textit{Thank you for your time, looks like I'm building it! c:};
\textit{Thank you for your help!};
\textit{Oh, I'd rather get a new case then. [...] Should I just throw it away?}
&
I'm glad you found it. It's easy to make a mistake, like Reddit isn't graded! :D \\
\midrule
\textbf{Target author B:}
\textit{My two cents: If you are asking technical questions then yes definitely. [...]};
\textit{Hey someone photoshopped out the tea label!};
\textit{I bet that the bot power is conducted through the floor.};
\textit{Or poke around on the ground};
\textit{That sucks. I thought dog treats were like essential kit [...]}
&
Nice. Easy to make the mistake. Like reddit isn't graded or anything. \\
\midrule[0.8pt]

\addlinespace[3pt]
\multicolumn{2}{p{0.96\textwidth}}{\textbf{Source 3:} Approx 35 people attended, which meant there was more than enough beer to go around.} \\
\addlinespace[2pt]
\midrule
\textbf{Target author A:}
\textit{and that's a record .. i had a temporary vacation from blogging because of my preliminary exams earlier this week!!};
\textit{i didnt take it badly though.. because he reasoned out that there was traffic along e. rod or something...};
\textit{i was sarcastic.. coz i was reviewing.. [...] luv u!!!};
\textit{if there's one way i would like to express my thoughts and ideas.. it is through writing!!};
\textit{ahhh.. naaalala ko na!!}
&
yep.. there was more than enough beer to go around, because 35 people attended!! \\
\midrule
\textbf{Target author B:}
\textit{But they're having a party on the 17th - not massive, only 50 people - so everybody can come over and eat lots.};
\textit{I'll keep ya looped.};
\textit{Well, I'm free for 10 whole weeks.};
\textit{Met some new people.};
\textit{Well, I've been on Prophet's Inc. and Fantasy Essentials - my favourite forums - for the last few hours, chatting to random people.}
&
It was a good night-- 35 people attended and there was more than enough beer to go around. \\
\bottomrule
\end{tabular}
\caption{Illustrative examples showing that HyperStyler generates different outputs depending on the target-author references given the same source text.}
\label{tab:qualitative_target_dependent}
\end{table*}

\begin{table*}[t]
\centering
\scriptsize
\setlength{\tabcolsep}{4pt}
\renewcommand{\arraystretch}{1.2}
\begin{tabular}{c p{0.37\textwidth} p{0.56\textwidth}}
\hline
\textbf{Cluster} & \textbf{Stylistic description} & \textbf{Representative examples} \\
\hline
\addlinespace[4pt]

1 &
\textbf{First-person informal narrative reply.}
These instances are characterized by first-person narration, informal register, and anecdotal narrative structure. &
\begin{tabular}[t]{@{}l@{}}
``I certainly did! [...] I was 37 at the time [...] haha'' \\
``That was my sibling, haha! Siblings are cruel ;\_;'' \\
``As I was going through your list, I was thinking `Ha! I know
all these guys!' [...]''
\end{tabular}
\\
\addlinespace[4pt]

2 &
\textbf{Brief appreciative reaction with emoticons.}
These instances are short acknowledgment responses marked by appreciation, positive evaluation, and emoticon usage. &
\begin{tabular}[t]{@{}l@{}}
``It did indeed work, thank you so much :D'' \\
``I actually lost the little plastic bit on the end, but I definitely will glue it back in, thanks :)'' \\
``I have no funny / awesome screenshots to share, but hey, maybe
if I win I can take some ;D''
\end{tabular}
\\
\addlinespace[4pt]

6 &
\textbf{Terse information-seeking question.}
These instances are short interrogative utterances that function primarily as follow-up requests for information. &
\begin{tabular}[t]{@{}l@{}}
``What about the baby versions?'' \\
``When did you get it?'' \\
``What happens when you break it?''
\end{tabular}
\\
\addlinespace[4pt]

11 &
\textbf{Extended clause-heavy justification reply.}
These instances are characterized by long, multi-clause constructions that build toward a conclusion through chained reasoning, conditional framing, or supporting elaboration. &
\begin{tabular}[t]{@{}l@{}}
``There is virtually no chance of getting them both. If it were
possible, of course \\ it's a great strategy [...]'' \\
``For sure. People look more into where a player was drafted than their actual skill [...]'' \\
``The fact that the handle is his name has a lot to do with it [...]''
\end{tabular}
\\
\addlinespace[4pt]

14 &
\textbf{Extended conversational reply.}
These instances are long, chatty responses combining informal register, colloquial markers, and loosely connected clauses. &
\begin{tabular}[t]{@{}l@{}}
``I love the song a lot haha. You should check out [...]'' \\
``Lol they were out of stock pretty much as soon as that price was up. [...]'' \\
``Yeah I was so hype when Honedge was announced [...]''
\end{tabular}
\\
\addlinespace[4pt]

15 &
\textbf{Formulaic endorsement.}
These instances are short evaluative formulas used to express approval, endorsement, or visibility boosting. &
\begin{tabular}[t]{@{}l@{}}
``Upvote for the title!'' \\
``This needs to be higher up in the comments!'' \\
``Well worth the hike!''
\end{tabular}
\\
\addlinespace[4pt]

16 &
\textbf{Directive second-person reply.}
These instances express second-person guidance through imperative constructions or modal advisory forms directed at the interlocutor. &
\begin{tabular}[t]{@{}l@{}}
``Google a video on how to make it.'' \\
``You should put her on the side then.'' \\
``You should post pictures sometime of the village if you have any.''
\end{tabular}
\\
\addlinespace[4pt]

18 &
\textbf{Segmented multi-line commentary.}
These instances are structured as short line-separated discourse units, often juxtaposing multiple evaluative or explanatory statements. &
\begin{tabular}[t]{@{}l@{}}
``A good landing is one where no-one gets hurt. [...]'' \\
``Symbols are dangerous. I love my heritage. [...]'' \\
``Wow. [...] The type of video that just leaves you speechless. [...]''
\end{tabular}
\\
\addlinespace[4pt]

23 &
\textbf{Blunt categorical assertion.}
These instances express direct and compact judgments or corrections in forceful declarative form. &
\begin{tabular}[t]{@{}l@{}}
``It's blue because he's cold'' \\
``If all that's on your resume you'll be fine'' \\
``If it's just the glass on top it's cheap''
\end{tabular}
\\
\addlinespace[4pt]

30 &
\textbf{Quote-and-correct reply structure.}
These instances exhibit a quote-response structure in which quoted content is followed by contradiction, correction, or practical follow-up. &
\begin{tabular}[t]{@{}l@{}}
``> Have had our house professionally cleaned and it still smells [...] \\ Wash the walls with diluted vinegar.'' \\
``> and that's a man with no engineering or mechanical educational background [...] \\ Education is completely irrelevant here.'' \\
``> bring em and register [...] There is no legal obligation to
register your firearms [...]''
\end{tabular}
\\
\addlinespace[4pt]

32 &
\textbf{Rhetorical second-person interrogative reply.}
These instances take interrogative form with ironic or sarcastic phrasing directed at the interlocutor, functioning as implicit challenge rather than genuine information-seeking. &
\begin{tabular}[t]{@{}l@{}}
``Don't you think that's setting the bar a little high?'' \\
``You sure you aren't just hearing Hanley's bat?'' \\
``Are you sure your teacher isn't Dwight Schrute?''
\end{tabular}
\\
\addlinespace[4pt]

41 &
\textbf{Punctuation-heavy expressive reply.}
These instances are characterized by repeated or emphatic punctuation and expressive phrasing, often conveying heightened emotional intensity. &
\begin{tabular}[t]{@{}l@{}}
``it is my top 2 as well!! fantastic story, great visuals and 
amazing characters!!'' \\
``i was lucky enough to be there! [...] and it was FANTASTIC!!!'' \\
``we will make it and have a roo meet up [...] BEAT THE CAPS!!''
\end{tabular}
\\
\addlinespace[4pt]

45 &
\textbf{Emphatic congratulatory/supportive reaction.}
These instances convey overt positive evaluation through repeated exclamation marks, congratulatory formulas, encouragement, and emoticons. &
\begin{tabular}[t]{@{}l@{}}
``Amazing!! Really beautiful gift :)'' \\
``You can do it!! Good luck =)'' \\
``Awww!! So sweet!! :) Congrats on the wedding!!''
\end{tabular}
\\
\addlinespace[4pt]

47 &
\textbf{Ellipsis-heavy hesitant reply.}
These instances are marked by repeated ellipses and loosely connected clauses, producing a hesitant and trailing conversational rhythm. &
\begin{tabular}[t]{@{}l@{}}
``that... that man has had some BAD food...'' \\
``[...] our humanity and kindness can circumvent that...'' \\
``this... this perfectly describes the whole datamining standpoint...''
\end{tabular}
\\
\addlinespace[4pt]

48 &
\textbf{Extended formal expository prose.}
These instances are characterized by long-form prose with formal register, technical or domain-specific vocabulary, and dense information structure. &
\begin{tabular}[t]{@{}l@{}}
``The events are in-game ones which cause unusual monster spawns, unusual numbers \\ of monsters, or give monsters new abilities.'' \\
``Traditional IRA contributions are an above-the-line deduction 
that happens \\ before itemized or standard deductions [...]'' \\
``Finding the minimum/maximum is all about finding out the points at which \\ the derivative of the function is 0 [...]''
\end{tabular}
\\
\addlinespace[4pt]
\hline
\end{tabular}
\caption{Selected clusters from a k-means clustering (k=50) in the STYLE embedding space. Sentences in each cluster group share distinct stylistic characteristics, illustrating that the STYLE embedder captures interpretable stylistic patterns independently of topic and content.}
\label{tab:cluster_style_summary}
\end{table*}

\begin{table*}[]
\centering
\resizebox{\textwidth}{!}{%
\begin{tabular}{@{}lcccc|cccc|cccc@{}}
\toprule
\multirow{2}{*}{Method} & \multicolumn{4}{c|}{\textbf{Reddit}} & \multicolumn{4}{c|}{\textbf{Blog}} & \multicolumn{4}{c}{\textbf{News}} \\
 & \textsc{Away} & \textsc{Towards} & \textsc{Sim} & \textsc{Joint} & \textsc{Away} & \textsc{Towards} & \textsc{Sim} & \textsc{Joint} & \textsc{Away} & \textsc{Towards} & \textsc{Sim} & \textsc{Joint} \\ \midrule
ASTRAPOP & 0.578 & 0.027 & 0.728 & 0.171 & 0.997 & 0.255 & 0.014 & 0.060 & 0.840 & 0.060 & 0.170 & 0.139 \\
ASTRAPOP$_{\text{JOINT}}$ & 0.620 & 0.029 & 0.695 & 0.173 & 0.840 & 0.070 & 0.171 & 0.139 & 0.576 & 0.082 & 0.713 & 0.319 \\
ASTRAPOP$_{\text{reverse}}$ & 0.571 & 0.026 & 0.743 & 0.172 & 0.991 & 0.275 & 0.077 & 0.170 & 0.655 & 0.086 & 0.394 & 0.248 \\
ASTRAPOP$_{\text{reverse,JOINT}}$ & 0.596 & 0.026 & 0.725 & 0.172 & 0.988 & 0.243 & 0.125 & 0.213 & 0.950 & 0.005 & 0.202 & 0.115 \\ \bottomrule
\end{tabular}%
}
\caption{Performance comparison results on the configuration of ASTRAPOP.}
\label{table12}
\end{table*}

\begin{table*}[]
\centering
\resizebox{\textwidth}{!}{%
\begin{tabular}{@{}lcccc|cccc|cccc@{}}
\toprule
\multirow{2}{*}{Method} & \multicolumn{4}{c|}{\textbf{Reddit}} & \multicolumn{4}{c|}{\textbf{Blog}} & \multicolumn{4}{c}{\textbf{News}} \\
 & \textsc{Away} & \textsc{Towards} & \textsc{Sim} & \textsc{Joint} & \textsc{Away} & \textsc{Towards} & \textsc{Sim} & \textsc{Joint} & \textsc{Away} & \textsc{Towards} & \textsc{Sim} & \textsc{Joint} \\ \midrule
ASTRAPOP (LLaMA-2-7B) & 0.708 & 0.188 & 0.505 & 0.333 & 0.813 & 0.244 & 0.569 & 0.479 & 0.799 & 0.110 & 0.559 & 0.322 \\
HyperStyler (T5-large) & 0.818 & 0.152 & 0.578 & \multicolumn{1}{c|}{\textbf{0.418}} & 0.731 & 0.183 & 0.701 & \multicolumn{1}{c|}{\textbf{0.489}} & 0.571 & 0.098 & 0.678 & \textbf{0.370} \\
\bottomrule
\end{tabular}%

}
\caption{Comparison between ASTRAPOP with its original backbone and HyperStyler.}
\label{ASTRAPOP_LLaMA}
\end{table*}

\begin{table*}[]
\centering
\resizebox{\textwidth}{!}{%
\begin{tabular}{@{}lcccc|cccc|cccc@{}}
\toprule
\multirow{2}{*}{Method} & \multicolumn{4}{c|}{Diverse} & \multicolumn{4}{c|}{Random} & \multicolumn{4}{c}{Single} \\
 & $\textsc{Away}$ & $\textsc{Towards}$ & $\textsc{Sim}$ & $\textsc{Joint}$ & $\textsc{Away}$ & $\textsc{Towards}$ & $\textsc{Sim}$ & $\textsc{Joint}$ & $\textsc{Away}$ & $\textsc{Towards}$ & $\textsc{Sim}$ & $\textsc{Joint}$ \\ \midrule
STYLL(Qwen2.5-7B) & 0.797 & 0.069 & 0.405 & 0.200 & 0.739 & 0.074 & 0.416 & 0.240 & 0.896 & 0.047 & 0.463 & 0.184 \\
gpt-4-turbo-2024-04-09 & 0.832 & 0.073 & 0.683 & 0.296 & 0.760 & 0.087 & 0.706 & 0.332 & 0.850 & 0.083 & 0.715 & 0.313 \\
gpt-5-mini-2025-08-07 & 0.861 & 0.082 & 0.718 &0.321 & 0.805 & 0.080 & 0.736 & 0.330 & 0.902 & 0.083 & 0.729 & 0.346 \\
gpt-5.4-2026-03-05 (medium) & 0.940 & 0.077 & 0.501 & 0.238 & 0.866 & 0.139 & 0.679 & 0.436 & 0.948 & 0.135 & 0.610 & 0.402 \\
meta-llama/Llama-3.1-8B-Instruct & 0.724 & 0.119 & 0.602 & 0.371 & 0.743 & 0.127 & 0.564 & 0.369 & 0.800 & 0.158 & 0.595 & 0.431 \\ \midrule
ParaGuide$_{\lambda=200}$ & 0.774 & 0.056 & 0.544 & 0.221 & 0.696 & 0.047 & 0.585 & 0.222 & 0.818 & 0.057 & 0.664 & 0.263 \\
ParaGuide$_{\lambda=2500}$ & 0.859 & 0.078 & 0.381 & 0.239 & 0.801 & 0.065 & 0.456 & 0.246 & 0.900 & 0.058 & 0.512 & 0.235 \\
StyleMC & 0.603 & 0.051 & 0.462 & 0.189 & 0.625 & 0.039 & 0.453 & 0.173 & 0.746 & 0.017 & 0.435 & 0.100 \\ \midrule
ASTRAPOP & 0.612 & 0.031 & 0.679 & 0.168 & 0.516 & 0.027 & 0.736 & 0.192 & 0.607 & 0.022 & 0.770 & 0.152 \\
ASTRAPOP$_{\text{JOINT}}$ & 0.656 & 0.035 & 0.645 & 0.168 & 0.550 & 0.029 & 0.702 & 0.205 & 0.653 & 0.022 & 0.739 & 0.145 \\ \midrule
TinyStyler$_{\text{REC}}$ & 0.897 & 0.130 & 0.306 & 0.282 & 0.863 & 0.154 & 0.314 & 0.315 & 0.932 & 0.148 & 0.436 & 0.371 \\
TinyStyler$_{\text{REC,RERANK(5)}}$ & 0.883 & 0.127 & 0.462 & 0.346 & 0.854 & 0.148 & 0.465 & 0.384 & 0.927 & 0.149 & 0.592 & 0.432 \\
TinyStyler & 0.836 & 0.109 & 0.582 & 0.356 & 0.835 & 0.127 & 0.590 & 0.396 & 0.910 & 0.130 & 0.706 & 0.445 \\
TinyStyler$_{\text{RERANK(5)}}$ & 0.831 & 0.111 & 0.693 & 0.393 & 0.839 & 0.125 & 0.705 & 0.434 & 0.907 & 0.131 & 0.793 & 0.480 \\ \midrule
HyperStyler$_{\text{REC}}$ & 0.810 & 0.162 & 0.426 & 0.368 & 0.748 & 0.163 & 0.468 & 0.384 & 0.857 & 0.143 & 0.538 & 0.399 \\
HyperStyler$_{\text{REC,RERANK(5)}}$ & 0.803 & 0.159 & 0.650 & 0.449 & 0.746 & 0.159 & 0.709 & 0.472 & 0.849 & 0.138 & 0.765 & 0.470 \\
HyperStyler & 0.813 & 0.153 & 0.547 & 0.402 & 0.768 & 0.157 & 0.542 & 0.410 & 0.872 & 0.145 & 0.645 & 0.443 \\
HyperStyler$_{\text{RERANK(5)}}$ & 0.807 & 0.149 & 0.760 & 0.467 & 0.766 & 0.148 & 0.768 & 0.480 & 0.871 & 0.145 & 0.844 & 0.508 \\ \bottomrule
\end{tabular}%
}
\caption{Performance comparison results on Reddit (Diverse / Random / Single) splits.}
\label{tab:my-table}
\end{table*}

\begin{table*}[]
\centering
\resizebox{\textwidth}{!}{%
\begin{tabular}{@{}llcccc|cccc|cccc@{}}
\toprule
\multicolumn{2}{l}{\multirow{2}{*}{Model}} & \multicolumn{4}{c|}{Diverse} & \multicolumn{4}{c|}{Random} & \multicolumn{4}{c}{Single} \\
\multicolumn{2}{l}{} & \textsc{Away} & \textsc{Towards} & \textsc{Sim} & \textsc{Joint} & \textsc{Away} & \textsc{Towards} & \textsc{Sim} & \textsc{Joint} & \textsc{Away} & \textsc{Towards} & \textsc{Sim} & \textsc{Joint} \\ \midrule
\multicolumn{2}{l}{HyperStyler} & 0.813 & \textbf{0.153} & 0.547 & \textbf{0.402} & 0.768 & \textbf{0.157} & 0.542 & \textbf{0.410} & 0.872 & \textbf{0.145} & 0.645 & \textbf{0.443} \\
 & \textit{w/o Stylo-navigator (Mean-pooling)} & 0.776 & 0.088 & 0.677 & 0.332 & 0.732 & 0.112 & 0.666 & 0.389 & 0.842 & 0.097 & 0.775 & 0.385 \\
 & \textit{w/o Stylo-navigator (Implicit selection)} & 0.770 & 0.102 & 0.641 & 0.348 & 0.729 & 0.124 & 0.627 & 0.392 & 0.833 & 0.116 & 0.746 & 0.412 \\
 & \textit{w/o Stylo-hypernet (Global)} & 0.985 & 0.009 & 0.142 & 0.015 & 0.986 & 0.008 & 0.150 & 0.033 & 0.999 & 0.000 & 0.202 & 0.002 \\
 & \textit{w/o Stylo-hypernet (Layer-wise)} & 0.789 & 0.119 & 0.597 & 0.374 & 0.734 & 0.123 & 0.595 & 0.388 & 0.852 & 0.121 & 0.697 & 0.421 \\
 & \textit{w/o adapter in FFN} & 0.794 & 0.129 & 0.579 & 0.385 & 0.747 & 0.139 & 0.565 & 0.400 & 0.859 & 0.134 & 0.681 & 0.443 \\
 & \textit{w/o prefix in CrossAttn} & 0.830 & 0.150 & 0.521 & 0.396 & 0.768 & 0.156 & 0.512 & 0.399 & 0.877 & 0.142 & 0.621 & 0.430 \\
 & \textit{w/ prefix in SelfAttn} & 0.958 & 0.022 & 0.416 & 0.094 & 0.960 & 0.016 & 0.434 & 0.081 & 0.990 & 0.010 & 0.529 & 0.071 \\
 & \textit{w/o predicted $z$ in stage 3 (mean-pooling)} & 0.809 & 0.143 & 0.542 & 0.399 & 0.768 & 0.157 & 0.546 & 0.407 & 0.878 & 0.133 & 0.654 & 0.421 \\
 & \textit{Underlying paraphraser} & 0.889 & 0.019 & 0.682 & 0.123 & 0.849 & 0.012 & 0.696 & 0.084 & 0.949 & 0.007 & 0.776 & 0.058 \\ \bottomrule
\end{tabular}%
}
\caption{Ablation study results on Reddit (Diverse / Random / Single) splits.}

\label{table:ablation_reddit}
\end{table*}

\begin{table*}[]
\footnotesize
\centering
\resizebox{\textwidth}{!}{%
\begin{tabular}{@{}cccccc|cccc|cccc@{}}
\toprule
\multirow{2}{*}{Rank} & \multirow{2}{*}{Prefix} & \multicolumn{4}{c|}{Diverse} & \multicolumn{4}{c|}{Random} & \multicolumn{4}{c}{Single} \\
 &  & $\textsc{Away}$ & $\textsc{Towards}$ & $\textsc{Sim}$ & $\textsc{Joint}$ & $\textsc{Away}$ & $\textsc{Towards}$ & $\textsc{Sim}$ & $\textsc{Joint}$ & $\textsc{Away}$ & $\textsc{Towards}$ & $\textsc{Sim}$ & $\textsc{Joint}$ \\ \midrule
8 & - & 0.814 & 0.141 & 0.546 & 0.396 & 0.760 & 0.149 & 0.531 & 0.400 & 0.870 & 0.141 & 0.641 & 0.438 \\
8 & 5 & 0.819 & 0.147 & 0.551 & 0.400 & 0.763 & 0.154 & 0.540 & 0.403 & 0.870 & 0.143 & 0.648 & 0.440 \\ \midrule
128 & - & 0.816 & 0.154 & 0.516 & 0.401 & 0.781 & 0.164 & 0.500 & 0.401 & 0.875 & 0.143 & 0.618 & 0.429 \\
128 & 5 & 0.822 & 0.156 & 0.524 & 0.403 & 0.770 & 0.164 & 0.528 & 0.411 & 0.877 & 0.148 & 0.627 & 0.440 \\ \midrule
- & 1 & 0.792 & 0.125 & 0.600 & 0.391 & 0.741 & 0.130 & 0.591 & 0.390 & 0.853 & 0.128 & 0.701 & 0.437 \\
32 & 1 & 0.818 & 0.145 & 0.538 & 0.396 & 0.764 & 0.150 & 0.536 & 0.397 & 0.871 & 0.145 & 0.640 & 0.438 \\ \midrule
- & 10 & 0.807 & 0.132 & 0.573 & 0.381 & 0.752 & 0.138 & 0.559 & 0.398 & 0.864 & 0.141 & 0.674 & 0.447 \\
32 & 10 & 0.817 & 0.149 & 0.528 & 0.399 & 0.772 & 0.163 & 0.523 & 0.409 & 0.877 & 0.147 & 0.632 & 0.438 \\ \bottomrule
\end{tabular}%
}
\caption{Hyperparameter study results on Reddit (Diverse / Random / Single) splits.}
\label{tab:Hyperparameter_study_results}
\end{table*}

\begin{table*}[t]
\centering
\small
\begin{tabular}{lllc}
\toprule
\textbf{Type} & \textbf{Artifact} & \textbf{License} & \textbf{Link} \\
\midrule
\multirow{5}{*}{Model}
 & UAR embedding & Apache-2.0 & \url{https://huggingface.co/rrivera1849/LUAR-MUD} \\
 & STYLE embedding & MIT & \url{https://huggingface.co/AnnaWegmann/Style-Embedding} \\
 & Paraphrasing PEGASUS & Apache-2.0 & \url{https://huggingface.co/tuner007/pegasus_paraphrase} \\
 & Mutual Implication Score & CC BY-NC-SA 4.0 & \url{https://github.com/s-nlp/mutual_implication_score} \\
 & T5-large & Apache-2.0 & \url{https://huggingface.co/google/t5-v1_1-large} \\
\midrule
\multirow{6}{*}{Software}
 & HuggingFace Transformers & Apache-2.0 & \url{https://github.com/huggingface/transformers} \\
 & Accelerate & Apache-2.0 & \url{https://github.com/huggingface/accelerate} \\
 & Scikit-learn & BSD-3-Clause & \url{https://scikit-learn.org/stable/} \\
 & NLTK & Apache-2.0 & \url{https://www.nltk.org/} \\
 & Matplotlib & PSF & \url{https://matplotlib.org/stable/project/license.html} \\
 & PyTorch & \href{https://github.com/pytorch/pytorch?tab=License-1-ov-file#readme}{License link} & \url{https://github.com/pytorch/pytorch} \\
\bottomrule
\end{tabular}
\caption{Used artifacts and their licenses and links. All artifacts are used consistent with their intended use.}
\label{tab:artifacts}
\end{table*}

\end{document}